\documentclass{article}
\usepackage{arxiv}
\usepackage[utf8]{inputenc}
\usepackage[T1]{fontenc}
\usepackage{times}
\usepackage{amsmath,amssymb,amsfonts}
\usepackage{booktabs}
\usepackage{graphicx}
\usepackage[dvipsnames]{xcolor}
\usepackage[hyphens]{url}
\usepackage[numbers,sort&compress]{natbib}
\usepackage[format=plain,labelformat=simple,labelsep=period,font=small]{caption}
\usepackage[font=footnotesize,skip=3pt,subrefformat=parens]{subcaption}
\usepackage[shortlabels,inline]{enumitem}
\usepackage{xspace}
\usepackage{microtype}

\usepackage{cuted} %

\usepackage{currfile} %

\usepackage{caption} %

\definecolor{cvprblue}{rgb}{0.21,0.49,0.74}
\usepackage[pagebackref,breaklinks,colorlinks,allcolors=cvprblue]{hyperref}
\usepackage[capitalize]{cleveref}
\crefname{section}{Sec.}{Secs.}
\Crefname{section}{Section}{Sections}
\Crefname{table}{Table}{Tables}
\crefname{table}{Tab.}{Tabs.}

\makeatletter
\DeclareRobustCommand\onedot{\futurelet\@let@token\@onedot}
\def\@onedot{\ifx\@let@token.\else.\null\fi\xspace}
\def\eg{\emph{e.g}\onedot} \def\ie{\emph{i.e}\onedot}

\makeatother
\setlist[itemize]{noitemsep,leftmargin=*,topsep=0em}

\newcommand{\thetitle}{MillBill: A Machining Agent with Geometric Feedback}
\def\maketitlesupplementary{\newpage
  \begin{center}{\Large\bfseries \thetitle}\\[0.5em]{\large Supplementary Material}\end{center}\vspace{1em}}

\title{\thetitle}
\renewcommand{\shorttitle}{MillBill: A Machining Agent with Geometric Feedback}

\author{Gennadii Savrasov$^{1,2,*}$ \quad Artem Slavin$^{1,*}$ \quad Soslan Kabisov$^{1}$\\
Andrey Kuznetsov$^{3,4}$ \quad Anton Konushin$^{1,2}$ \quad Dmitrii Zhemchuzhnikov$^{1,2,\dagger}$\\[0.6em]
{\normalfont $^{1}$Lomonosov Moscow State University \quad $^{2}$AXXX, Moscow, Russia \quad $^{3}$Innopolis University \quad $^{4}$FusionBrain Lab}\\[0.3em]
{\normalfont\small $^{*}$Equal contribution \quad $^{\dagger}$Corresponding author: \texttt{zhemchuzhnikovds@my.msu.ru}}}
\date{}

\begin{document}
\maketitle
\begin{abstract}
Turning a CAD model into a machining program requires both process knowledge and geometric precision. Language models can generate such workflows by writing geometry and toolpath code for each part, but this makes numerical execution part of the model's task. We introduce MillBill, an agent that separates machining decisions from their geometric execution and verification. Given a boundary-representation part, stock, and material, a locally hosted, mid-sized open model proposes operations and cutting parameters for indexed three-axis machining. A geometric harness generates toolpaths, simulates material removal, and returns feedback for plan refinement. Setup planning accounts for what the available cutters can physically reach, rather than visibility alone. We evaluate the resulting programs by the fraction of the target surface area within the specified tolerance after simulated machining. With Qwen3.8-27B, MillBill outperforms frontier-model approaches that construct their own geometry and toolpath workflows on the evaluated machining tasks. Code: \url{https://github.com/kulibinai/millbill}.
\end{abstract}

\section{Introduction}
\label{sec:intro}

A CNC machining program is the end of a long chain of decisions. From a CAD model of the part, a CAM programmer chooses the operations for each feature and their order, the cutters, and the speeds and feeds; chooses the sides from which to approach the stock; and generates and simulates the toolpaths. The first choices draw on process knowledge. The rest is three-dimensional geometric reasoning on the boundary representation (B-rep) of the part, and it must be precise: a toolpath off by a fraction of a millimetre gouges the part or leaves material behind. Computer-aided process planning (CAPP) has aimed to automate this chain for decades, yet in practice it still relies on human experts~\cite{xu2011capp,alwswasi2018acapp}.

Language models add process knowledge to this problem, but existing approaches either leave the geometry unchecked or hand it to the model. LLM process planners produce operation sequences, tools and cutting parameters~\cite{cappgpt,arkness,designtoplan} but stop before toolpaths, so their plans are never executed against the part. Other work has the model write G-code directly~\cite{gllm,sket2025gcode,lee2026gcode}, and CNCGEN learns operations and toolpaths from B-rep models at the cost of a large machining training set~\cite{cncgen}. Given a part, a frontier model can also write and run its own geometry and toolpath code; we call this a \emph{code-writing frontier agent} and use it as our main baseline. In this setting, coordinates, offsets and swept volumes become the model's task, and a single wrong number yields a program that runs but machines the wrong part.

We argue that machining decisions and their geometric execution call for different machinery. Choosing the operation for a feature, the cutter and how aggressively to cut is a matter of process knowledge, where a language model is useful. Generating toolpaths, predicting the material they remove and measuring the result are geometric computations done by algorithms; the model sets only operation-level parameters such as the tool, the finishing allowance and the depth of cut, and never computes coordinates, offsets or swept volumes. Relieved of this numerical work, a locally hosted, mid-sized open model should suffice, so proprietary parts need not be sent to an outside provider; our experiments test this.

We introduce MillBill, an agent built on this separation. Given a B-rep part, its stock and its material, MillBill recognises machining features with a graph neural network on the B-rep~\cite{aagnet} and chooses setups among the six axis-aligned orientations of indexed three-axis machining. The language model then revises a standard-practice reference of operations, tools and cutting parameters derived from the features, the tool library and a cutting-data model; its tools must exist in the library and its cutting data are clamped to the machine limits. A \emph{geometric harness} --- CAM toolpath generation, material-removal simulation and surface-deviation measurement --- turns the plan into G-code, simulates the cut and attributes deviations from the design to features and operations. It corrects what it can, for example by adding rest finishing, returns the remaining deviations to the model, and keeps the best of at most three builds.

Setups are chosen by what the available cutters can reach, not by visibility alone: a deep, narrow region can be visible yet out of reach once the holder is taken into account. Accessibility has been computed before~\cite{inui2018accessibility,nelaturi2015spatial}; MillBill screens the whole tool library with one height-map dilation per remaining pair of holder overhang and reach, after removing redundant pairs, picks setups greedily and plans each on the stock the simulated earlier setups leave.

We evaluate on 100 parts collected from GrabCAD and score every system by simulating its G-code. With the open Qwen3.8-27B model~\cite{qwen38}, MillBill leaves 97.4\% of the target surface within 0.1\,mm, against 94.3\% for the best code-writing agent built on a frontier model, GPT-6 Astra~\cite{gpt6astra}. Most of the gain comes from the harness: without the model, it already reaches 97.0\%.

Our contributions are:
\begin{itemize}
  \item MillBill, a B-rep-to-G-code pipeline in which a language model makes the machining decisions and a geometric harness executes and verifies them (\cref{sec:method});
  \item an open benchmark of 4{,}294 GrabCAD parts paired with milling stocks, with a protocol that scores any planner's G-code in simulation (\cref{sec:exp:setup});
  \item state-of-the-art results on this benchmark with a locally hosted open model, which outperforms code-writing agents built on frontier models (\cref{sec:exp}).
\end{itemize}

\section{Related Work}
\label{sec:related}

\paragraph{Machining feature recognition.}
Learned machining feature recognition moved from voxel grids~\cite{featurenet} to graphs on the B-rep~\cite{mfcad,hierarchicalcadnet}, alongside general B-rep encoders~\cite{uvnet,brepnet}. AAGNet segments a geometric attributed adjacency graph into feature classes, instances and bottom faces in one network and introduced the MFInstSeg dataset~\cite{aagnet}; later work uses transformers, domain adaptation, self-supervised pretraining or panoptic segmentation~\cite{brepformer,brepmfr,brepmae,featurefox}. Vision--language models prompted with renderings can also name features, but produce spurious labels~\cite{khan2025vlmfr}. Recognition is not our contribution: MillBill uses AAGNet for instance segmentation and measures feature dimensions on the B-rep.

\paragraph{Process planning, setups and simulation.}
Computer-aided process planning has a long history of rule-based, knowledge-based and optimisation-based systems~\cite{xu2011capp}, and fully automatic planning remains open~\cite{alwswasi2018acapp}. Setup planning is commonly posed through visibility, using visibility maps on the Gaussian sphere~\cite{woo1994visibility} or global visibility maps computed from slices of the part~\cite{hou2017gvm}. Accessibility cones that account for the tool and holder have been computed for 3+2-axis milling~\cite{inui2018accessibility}, configuration-space formulations derive the volume each tool can remove~\cite{nelaturi2015spatial,behandish2018hybrid}, and cutter selection has been posed as a trade-off between accessible area and machining time~\cite{yao2003cutters}. MillBill applies a holder-aware reach test for the whole tool library, one height-map dilation per remaining pair of holder overhang and reach, after removing redundant pairs, and selects indexed setups greedily by the material they newly make reachable rather than by visibility. To verify programs, material-removal simulation represents the workpiece with Z-maps~\cite{anderson1978zmap}, dexels~\cite{vanhook1986dexel,zhang2009tridexel} or voxels~\cite{joy2017voxel,schnoes2021gpuvoxel} and removes the swept volume of the tool~\cite{abdelmalek2006swept}; MillBill instead tests query points exactly against the analytically swept cutters, without discretising the workpiece, and uses dexel height maps only to localise the deviations it feeds back to the planner.

\paragraph{Language models for manufacturing and CAM.}
Language models have been surveyed and evaluated across manufacturing tasks~\cite{llm4manufacturing}. For process planning, CAPP-GPT generates process plans with a GPT-style framework~\cite{cappgpt} and ARKNESS grounds an LLM in machining knowledge graphs so that a small local model answers process-planning questions~\cite{arkness}. Design-to-Plan coordinates LLM agents that plan from 3D CAD models and 2D drawings~\cite{designtoplan}; like the planners above, it stops before toolpaths, so its plans are never executed or checked against the geometry. Other work has the model write machine code: GLLM generates G-code with a fine-tuned model and corrects it with validation feedback~\cite{gllm}, general-purpose chat models have been compared on writing G-code for a three-axis part~\cite{sket2025gcode}, and a neuro-symbolic approach constrains LLM G-code generation with separation-logic verification~\cite{lee2026gcode}; related studies have models debug G-code for additive manufacturing~\cite{jignasu2023gcode} or verify it against machine screens~\cite{hashempour2025vlmgcode}. CNCGEN, closest to our setting, learns operations and toolpaths from B-rep models on a large machining dataset~\cite{cncgen}; MillBill needs no training on machining plans or toolpaths: an off-the-shelf open model makes the decisions, a pretrained feature recogniser segments the part, and geometric algorithms execute and verify them. We do not compare with CNCGEN: it is trained on its own dataset, outputs operations and parameterised toolpaths rather than G-code, and, to our knowledge, its code and models are not publicly available.
CNC-Net learns without supervision to reproduce a shape with milling and drilling operations of cylindrical tools~\cite{cncnet}.

\paragraph{Tool-using agents with execution feedback.}
Language-model agents learn to call tools~\cite{toolformer}, interleave reasoning with actions~\cite{react}, and improve their outputs from feedback, whether self-generated or from execution~\cite{selfrefine,reflexion}. In CAD, learned models write CAD programs from point clouds and images~\cite{cadrille,cadevolve,cadena}, and agents generate and edit models through CAD kernels~\cite{cadassistant,itercad,toolcad,cadsmith,cadfather} and refine them from rendered or measured results~\cite{cadassistant,cadcodeverify,cadsmith,cadreasoner}. MillBill applies this pattern to machining, where the feedback is the deviation of the simulated part from the design rather than the appearance or validity of a model.

\section{MillBill}
\label{sec:method}

\paragraph{Problem.}
The input is a target part $\mathcal{P}$ given as a B-rep (STEP), a stock $\mathcal{S} \supseteq \mathcal{P}$, a workpiece material and, optionally, drawing notes such as threads, fits and surface-finish requirements. The stock is either a user-supplied blank or the bounding box of the part enlarged by fixed allowances. The machine is a three-axis vertical machining centre with a library $\mathcal{T}$ of cutters and a machine profile (spindle speed, feed, power and torque limits). The part may be machined in several indexed setups, each approaching it from one of the six axis directions $\mathcal{D} = \{\pm x, \pm y, \pm z\}$. The output is one G-code program per setup. A program is judged by the part it produces: after simulating all setups, we measure the fraction of the target surface that lies within a tolerance $\tau$ of the design (\cref{sec:method:verify}).

\begin{figure*}[t]
  \centering
  \includegraphics[width=\textwidth]{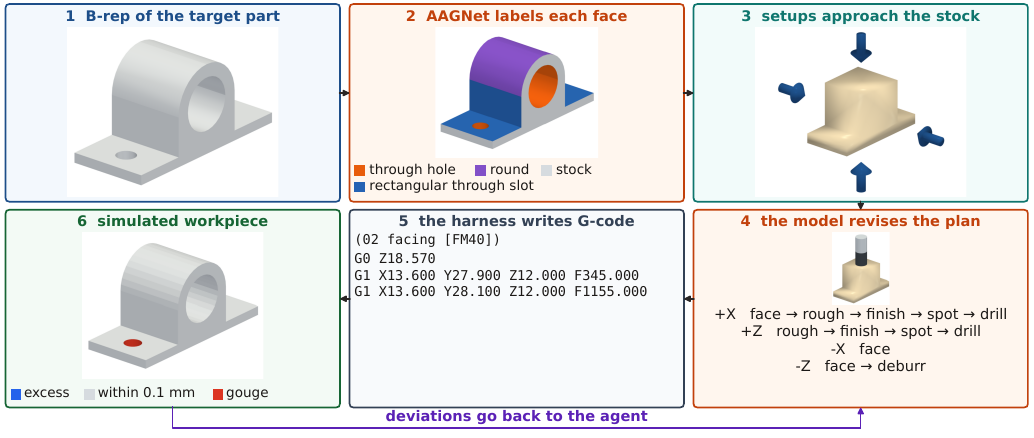}
  \caption{\textbf{MillBill overview.} The language model revises a standard-practice machining plan; the geometric harness executes it, simulates the result and returns its deviations from the design. (1)~The target B-rep. (2)~AAGNet labels each face with a machining feature class (\cref{sec:method:features}). (3)~Setups are chosen among the six axis directions by what the cutters can reach (\cref{sec:method:setups}). (4)~The model revises the operations, tools and cutting parameters of each setup, within the tool library and the machine limits (\cref{sec:method:agent}). (5)~The harness writes the toolpaths and G-code (\cref{sec:method:cam}). (6)~The simulated workpiece is compared with the design (blue: excess material, grey: within 0.1\,mm, red: gouge), and the deviations go back to the agent (\cref{sec:method:verify,sec:method:refine}).}
  \label{fig:overview}
\end{figure*}

\paragraph{Overview.}
MillBill splits the work between a language model and a geometric harness (\cref{fig:overview}). The harness recognises machining features on the B-rep (\cref{sec:method:features}) and chooses setups by tool reachability (\cref{sec:method:setups}). For each setup, the agent retrieves machining knowledge and decides the operation sequence, tools and cutting parameters (\cref{sec:method:agent}). The harness turns the plan into toolpaths and G-code (\cref{sec:method:cam}), simulates material removal and measures the result against the target (\cref{sec:method:verify}), and returns the deviations so that the plan can be refined (\cref{sec:method:refine}). The model never writes coordinates or G-code: every toolpath is generated by the harness, and every program is verified in simulation.

\subsection{Machining features}
\label{sec:method:features}

Machining decisions are made per feature, so the part is first segmented into machining features with AAGNet~\cite{aagnet}. AAGNet operates on a geometric attributed adjacency graph of the B-rep, whose nodes are faces and whose edges are shared B-rep edges, and predicts in one pass a semantic class for every face, a face-to-face affinity that groups faces into feature instances, and which faces form the bottom of a feature. We use the network trained on MFInstSeg~\cite{aagnet}, whose 24 classes cover holes, passages, slots, steps, pockets, chamfers, rounds and O-rings, plus a stock class. Faces predicted to belong to one feature are grouped, each face into at most one group; a group takes the class with the highest summed face scores, and stock groups are dropped.

The network determines which faces form a feature and what type it is; its dimensions are measured on the B-rep. AAGNet and FreeCAD load the STEP model independently, so their face indices may differ; each predicted face is matched to the FreeCAD face with the nearest centroid, within tolerances on centroid distance and area. For a hole, coaxial concave cylindrical faces give the diameter, the depth along the axis, whether it is through or blind, and the shape of its bottom; coaxial stacks become counterbores or countersinks. For pockets, slots and steps, the predicted bottom faces give the floor, from which we measure the floor height, depth, extent, minimum width and the smallest concave corner radius, and the class determines whether the feature is open to the side. Features whose floor faces away from the current setup are left to another setup.

Drawing notes are parsed by the language model into structured annotations (threads, tolerance grades, surface roughness, counts and locations), which are then matched to features. Each annotation--feature pair is scored as
\begin{equation}
  s = w_{\text{type}}\, s_{\text{type}} + w_{\text{dim}}\, s_{\text{dim}} + w_{\text{loc}}\, s_{\text{loc}},
  \label{eq:fusion}
\end{equation}
where $s_{\text{type}}$ checks that the annotation applies to the feature type, $s_{\text{dim}}$ compares annotated and measured dimensions (\eg, a thread matches a hole at its tap-drill or clearance diameter), and $s_{\text{loc}}$ matches spatial keywords such as ``top'' or ``blind''. Confident unique matches are accepted directly; ambiguous ones are resolved by the model, which may assign one annotation to several features. Matched annotations set how a feature is machined: a threaded hole gets a thread-cutting operation, and a hole tighter than drilling can hold is reamed or bored after drilling.

\subsection{Reach-aware setup planning}
\label{sec:method:setups}

A region of the part can be machined from direction $d$ only if a cutter can get there. Visibility from $d$ is necessary but not sufficient: a cutter has finite length and is clamped in a holder much wider than itself, so a deep, narrow region may be visible yet out of reach. We discretise the volume to be removed, $\mathcal{S} \setminus \mathcal{P}$, into voxels and, for each direction, call a voxel \emph{visible} if no part material lies above it along $d$. Let $H_d(x,y)$ be the height of the part's top surface in the frame of $d$. Each milling cutter $t \in \mathcal{T}$ with radius $r_t$ and usable length $L_t$ (overall length minus the clamped length), held in a holder of radius $r_h$, overhangs its own axis by $\rho_t = r_h - r_t$. A visible voxel at $(x, y, z)$ is \emph{reachable} from $d$ if some cutter clears the part around it:
\begin{equation}
  \exists\, t \in \mathcal{T}: \quad \max_{\|(x',y') - (x,y)\| \le \rho_t} H_d(x',y') \;-\; z \;\le\; L_t .
  \label{eq:reach}
\end{equation}
\Cref{eq:reach} is a necessary condition for holder clearance: it checks a disk of radius $\rho_t$ around the voxel, not the full holder around every admissible tool axis, so it can accept regions the holder would collide with. A cutter is skipped if another has at least the same usable length $L_t$ and no larger holder overhang $\rho_t$, since every voxel it reaches is also reached by the other. For each remaining $(\rho_t, L_t)$ pair, the maximum part height within a disk of radius $\rho_t$ around every height-map cell is computed once, as a grey-scale dilation of $H_d$, which tests \cref{eq:reach} for all voxels at once. Holes are assigned by visibility along the drilling direction, assuming sufficient drill reach; drill diameter and length are checked during tool selection.

Setups are chosen greedily. The primary setup is the direction that reaches the largest volume of material to be removed, with a preference for $+z$ when it is nearly as good. Further directions are added in order of the reachable volume they newly expose, until the gain falls below a small threshold or all six directions are used. A feature is assigned to a setup if most of its volume is reachable from it and not yet removed by earlier setups; features that no direction reaches are reported, and so is the unreachable volume. Each later setup is planned on the stock that the earlier setups actually leave, as measured by the simulation of \cref{sec:method:verify}.

\subsection{Machining decisions by the agent}
\label{sec:method:agent}

A language model revises the machining plan in three stages, following the agent architecture of Design-to-Plan~\cite{designtoplan}, each a ReAct agent~\cite{react} that alternates reasoning with calls to harness tools and ends with a structured JSON answer.
Knowledge retrieval runs once per part. For each setup, the harness prepares an initial plan: a standard-practice operation sequence, the cutters that pass geometric and material checks, and cutting data from a physical model. The process-sequence agent revises the operations and their order, and the tool-selection agent the tools and cutting parameters.

\paragraph{Knowledge retrieval.}
From the recognised features, their dimensions and the workpiece material, the first agent retrieves machining limits and recommendations, such as minimum internal corner radii, maximum hole depth-to-diameter ratios and material-specific cutting guidance, and flags violations (conflicts with these guidelines, not simulated defects); \eg, a hole deeper than the recommended ratio is flagged for the planning agents when they choose the drilling process and tool. Three sub-agents search textual guidelines, rule and thread tables, feature--material tables, a process knowledge graph and a materials database in parallel; their results are merged, keeping the highest severity of a conflict, into rules, violations and recommended processes for the other two agents.

\paragraph{Process sequence.}
The second agent decides which operations machine each feature and in what order. It sees the features, the matched annotations, the retrieved violations and recommendations, and a reference sequence that follows standard shop practice: datum faces first, then roughing of all features, finishing, hole making from spot drilling through drilling, chamfering or countersinking, reaming or boring to tapping, and deburring last. Its tools look up process capabilities, retrieve sequencing rules and validate a candidate sequence against hard precedence constraints (paired operations on the same feature, nested features, roughing before finishing). The proposed sequence is normalised to the process catalogue, minimally repaired by moving only the steps that violate a hard constraint, and accepted if it has no more critical violations than the reference.

\paragraph{Tools and cutting parameters.}
The third agent assigns a cutter and cutting parameters to every operation. Candidate cutters are filtered in four stages: whether the process needs a tool, which tool categories serve it, which cutters are rated for the material's ISO~513 group and fit the feature geometry (\eg, diameter below the slot width or corner radius, sufficient reach and flute length, drill diameter and length, tap thread series), and a ranking that prefers the largest cutter that fits. Reference cutting data come from a physical model: cutting speed from the material group and tool material, feed per tooth proportional to the cutter diameter with chip-thinning compensation at small radial engagement, depths of cut by strategy, and spindle power from the Kienzle force model, limited by the machine~\cite{kienzle1952}. The agent can change the tool, which must exist in the library with valid cutting data, and can adjust spindle speed, feed and depths of cut; speed and feed are clamped to the machine limits. Operations of the same tool within a phase are then grouped to avoid tool changes, provided no precedence constraint is broken.

\subsection{Toolpaths and G-code}
\label{sec:method:cam}

The harness executes the plan in the CAM workbench of FreeCAD~\cite{freecad}. Each setup becomes a CAM job on the part rotated so that the setup direction points up, with its stock and one tool controller per combination of cutter and cutting parameters. Facing, adaptive clearing of suitable closed pockets, outer profiling and drilling cycles use FreeCAD operations. Slots, steps, other pockets, cavity roughing and surface finishing use toolpaths computed by the harness on a height map of the target seen from the setup direction. Roughing paths keep the tool centre above the target dilated by the tool disk, which makes them gouge-free for flat cutters by construction, and proceed in z levels at the chosen depth of cut. Finishing paths follow waterlines obtained from an exact Euclidean distance transform of the material at each level~\cite{park2004zlevel}, with a raster pass on sloped regions. Counterbores and thread milling use helical paths. Every path computed by the harness passes through a gouge check against a dexel height map of the target, and any piece of a move that would cut into the target is lifted. FreeCAD's post-processor writes the G-code; canned cycles that are unsafe in its generic output are rewritten, and a safety check flags unsafe moves (\cref{sec:suppl:impl}).

\subsection{Simulation and verification}
\label{sec:method:verify}

The harness simulates every program without discretising the workpiece: it keeps only the list of cutter moves and answers, for any query point, whether material remains there. Each G0 or G1 line is a straight move of the cutter tip from $\mathbf{c}_0$ to $\mathbf{c}_1$; arcs are replaced by chords of bounded sagitta, and rapid moves cut like feed moves, so that a rapid into material is detected. A cutter is a solid of revolution with an unbounded shank, described by its profile $\eta(d)$, the height of the cutting edge above the tip at distance $d \le R$ from the axis: zero for flat end mills, a circular arc for ball and bull-nose end mills, and a cone for drills, spot drills, chamfer mills and countersinks. A move $m$ removes a point $\mathbf{p}$ if the cutter contains $\mathbf{p}$ at some position along the move,
\begin{equation}
  \exists\, t \in [0,1]:\;\; d(t) \le R \;\wedge\; p_z \ge c_z(t) + \eta\bigl(d(t)\bigr),
  \label{eq:removes}
\end{equation}
where $\mathbf{c}(t) = \mathbf{c}_0 + t(\mathbf{c}_1 - \mathbf{c}_0)$ and $d(t)$ is the planar distance from $\mathbf{p}$ to the cutter axis. The interval of $t$ with $d(t) \le R$ is given by a quadratic equation. On it, vertical and horizontal moves and inclined moves of flat end mills have a closed-form answer; for an inclined move of a profiled cutter, the edge height above $\mathbf{p}$, $c_z(t) + \eta(d(t))$, is convex in $t$ and its minimum is found by golden-section search. This is an exact test against the swept volume of the cutter~\cite{abdelmalek2006swept}, rather than an approximation on dexels~\cite{vanhook1986dexel} or voxels. After all setups, the workpiece is
\begin{equation}
  \mathcal{W} = \bigl\{\, \mathbf{p} \in \mathcal{S} : \nexists\, k,\, m \in \mathcal{M}_k \text{ removing } R_k\mathbf{p} + \mathbf{t}_k \,\bigr\},
  \label{eq:workpiece}
\end{equation}
where $\mathcal{M}_k$ are the moves of setup $k$ and $(R_k, \mathbf{t}_k)$ maps the part frame to its frame. Moves are binned by their planar footprint and sorted by height, so that each query inspects only the few moves that can reach the point.

We sample $N$ points $\mathbf{p}_i$ on the faces of the target B-rep, projected onto the exact surface, with exact outward normals $\mathbf{n}_i$ and area weights $a_i$. Each point is probed along its normal at the distances $r \in \rho_\tau = \{\tau/4, \tau/2, 3\tau/4, \tau\}$. The point has \emph{excess} material if $\mathbf{p}_i + r\mathbf{n}_i \in \mathcal{W}$ for every $r \in \rho_\tau$, and is \emph{gouged} if $\mathbf{p}_i - r\mathbf{n}_i \notin \mathcal{W}$ for every $r \in \rho_\tau$. The score is the fraction of the target surface area that has neither,
\begin{equation}
  \text{Acc}_\tau = \frac{\sum_i a_i\, [\, \neg\,\text{excess}_i \;\wedge\; \neg\,\text{gouge}_i \,]}{\sum_i a_i}.
  \label{eq:metric}
\end{equation}
The same function scores MillBill and the baselines and ranks the builds of \cref{sec:method:refine}. To localise errors for the feedback, the harness additionally keeps a dexel height map of each setup~\cite{vanhook1986dexel,anderson1978zmap}: connected regions of out-of-tolerance excess or gouge are attributed to the feature they lie on and to the last operation that cut there. Regions that no library cutter can resolve, such as sharp internal corners below the smallest cutter radius or depths beyond the longest reach, are labelled as inherent and are not fed back as errors.

\subsection{Refinement from geometric feedback}
\label{sec:method:refine}

After each build, the harness ranks the result by safety issues, gouged area and the score of \cref{eq:metric}, and keeps the best build seen so far. If the result is out of tolerance, the harness first adds rest finishing: it locates material left above the target, selects up to two library cutters that can reach and finish it, and machines it with parallel back-and-forth passes spaced so that the ridges left between passes are at most $\tau/4$ high. Defects that have a direct correction, such as an empty toolpath or a finishing allowance with no finishing pass, are corrected by the harness. The remaining defects are returned to the model together with the operations, the features and the attributed error regions, and the model proposes a minimal set of fixes from a fixed action space: switch the tool of an operation, set the finishing allowance, the final depth, the depth of cut, the stepover or the peck depth, or drop a finishing, deburring or spot-drilling operation. Each fix is validated (\eg, a switched tool must exist in the library with valid cutting data, a finishing allowance must lie in 0--1\,mm) but not bounded relative to the reference. The plan is rebuilt and re-verified, for at most three builds, and the loop stops early when a build is within tolerance or no longer improves.

\section{Experiments}
\label{sec:exp}

\subsection{Setup}
\label{sec:exp:setup}

\begin{figure}[t]
  \centering
  \includegraphics[width=0.4\linewidth]{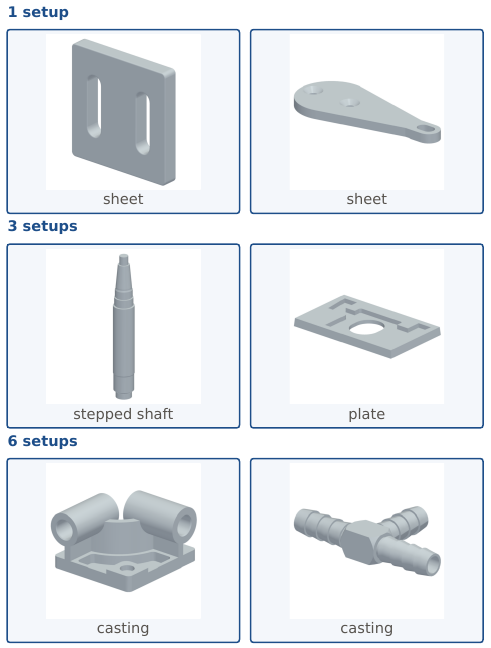}
  \caption{\textbf{Evaluation parts.} Six of the 100 parts, grouped by the number of setups and labelled with their stock kind.}
  \label{fig:parts}
\end{figure}

\paragraph{Parts.}
We evaluate on 100 parts drawn from a collection of 4{,}294 models selected at random from the GrabCAD Community Library\footnote{\url{https://grabcad.com/library}}, used under the GrabCAD Terms of Use (\cref{sec:suppl:data}). The collection pairs every single-solid STEP part with a milling stock. For 1{,}396 parts the stock is turned: 694 stepped shafts with 1\,mm of allowance, and 702 solids of revolution with no allowance, their local cavities filled out to the circumscribed circle. The other 2{,}898 stocks are prismatic: a stepped casting, forging or profile, or a rolled flat bar, square bar, plate, plate strip or sheet (\cref{sec:suppl:data}). The 100 parts are a stratified draw from this collection (random seed 2026): ten parts with a turned stock and ten from each of the eight prismatic stock kinds, then ten more from the remaining candidates. A candidate is one solid with 25 to 200 B-rep faces, at most 0.5\% of the removal volume that no three-axis cutter reaches from any of the six axis-aligned directions, and a workpiece material from the knowledge base (\cref{sec:suppl:data}); near-duplicate shapes are dropped, which removes 112 candidates. The draw yields 10 to 13 parts of each kind. Each task consists of the part as a STEP B-rep, the stock paired with it in the collection and the workpiece material (\cref{sec:method}). The parts range from 5.08 to 1{,}840\,mm along the longest bounding-box edge and from 25 to 179 B-rep faces, 90 of them require more than one setup (\cref{fig:parts}). Six parts are longer than the travel of a typical three-axis machining centre ($600 \times 400 \times 450$\,mm); we keep them, since the harness does not check machine travel. Their scores are in \cref{sec:suppl:geometry}. Unless stated otherwise, the tolerance is $\tau = 0.1$\,mm and all systems use the machine profile and tool library of \cref{tab:suppl:params}.

\paragraph{Baselines.}
\emph{Code-writing frontier agents} receive the same task as MillBill: the part and stock as STEP files, the material, the notes, the tolerance, the machine profile, the coordinate convention and the tool library, and must return one G-code program per setup together with a description of its tools and operations (\cref{sec:suppl:baseline}). The model reads the STEP files and writes the G-code itself: it can run Python, view images and read and write files, and so write and execute its own geometry and toolpath code. We build such agents on GPT-6 Astra~\cite{gpt6astra} and Claude Opus 5.5~\cite{opus55}, each with a budget of \$2 and at most 80 turns per part and no feedback on its programs, and on Qwen3.8-27B~\cite{qwen38}, the same model as in MillBill, which runs locally with at most 160 calls and four hours per part and receives two rounds of feedback from our simulator (\cref{sec:suppl:baseline}).
CNC-Net~\cite{cncnet} learns, without supervision on machining sequences, to predict milling and drilling operations with cylindrical tools that reproduce a given shape. Its published code simulates the material removed by the predicted operations and saves the result as a mesh; it does not convert the operations to G-code and does not accept a given stock. We therefore score this mesh, machined from the bounding box of the part. It is trained from scratch for each part.

\paragraph{Metrics.}
All G-code programs are simulated and scored by the same code (\cref{sec:method:verify}), and the CNC-Net meshes by the same metric. The primary metric is $\text{Acc}_\tau$ of \cref{eq:metric}, the fraction of target surface area within tolerance, averaged over the scored parts; we also report the areas with excess material and with gouges, which $\text{Acc}_\tau$ does not separate. A gouge cannot be repaired, while excess material can be removed by an additional pass, so the two errors are not equivalent.

\paragraph{Implementation.}
MillBill uses Qwen3.8-27B~\cite{qwen38} in FP8 on four local NVIDIA H100 80\,GB GPUs, with CAM and simulation on CPUs (\cref{sec:suppl:impl}). Each part gets one run of at most three hours and takes 16.1 minutes on average, 12.1 of them in CAM generation and its checks and 1.4 in planning.

\subsection{Main results}
\label{sec:exp:main}

\begin{table}[t]
  \centering
  \small
  \setlength{\tabcolsep}{4pt}
  \begin{tabular}{@{}lcccc@{}}
    \toprule
    Method & Scored & $\text{Acc}_\tau$ $\uparrow$ & Gouge $\downarrow$ & Excess $\downarrow$ \\
    \midrule
    CNC-Net~\cite{cncnet} & 99/100 & 42.3 & 31.11 & 26.6 \\
    \multicolumn{5}{@{}l}{\emph{Code-writing agent}} \\
    \quad Qwen3.8-27B & 82/100 & 7.9 & 38.99 & 53.5 \\
    \quad Opus 5.5 & 100/100 & 83.0 & 4.76 & 12.3 \\
    \quad GPT-6 Astra & 99/100 & 94.3 & 1.50 & 4.2 \\
    \midrule
    MillBill (ours) & 100/100 & \textbf{97.4} & \textbf{0.54} & \textbf{2.1} \\
    \bottomrule
  \end{tabular}
  \caption{\textbf{Comparison on the 100-part benchmark} at $\tau = 0.1$\,mm. Scored: parts scored successfully; failures are reported separately without imputed scores. $\text{Acc}_\tau$, gouge and excess: percentage of the target surface area of the final workpiece, averaged over the scored parts. CNC-Net uses its native bounding-box stock. Initial-stock scores, volume proxies, depths and band diagnostics are in \cref{tab:suppl:geometry}.}
  \label{tab:geometry-comparison}
\end{table}

\Cref{tab:geometry-comparison} compares MillBill with the baselines. MillBill leaves 97.4\% of the target surface within tolerance, against 94.3\% for the GPT-6 Astra agent, 83.0\% for the Opus 5.5 agent and 42.3\% for CNC-Net, and has the smallest gouged area (0.54\%, against 1.50\% for the best baseline) and the smallest area with excess material (2.1\%). The code-writing agents err mainly in the transformations of the $\pm x$ and $\pm y$ setups and in tool offsets, and some add deburring passes that cut into the part. As a code-writing agent, Qwen3.8-27B reaches 7.9\% on 82 parts, below the uncut stock (10.6\%), against 97.4\% inside MillBill.

\begin{figure*}[t]
  \centering
  \includegraphics[width=0.75\textwidth]{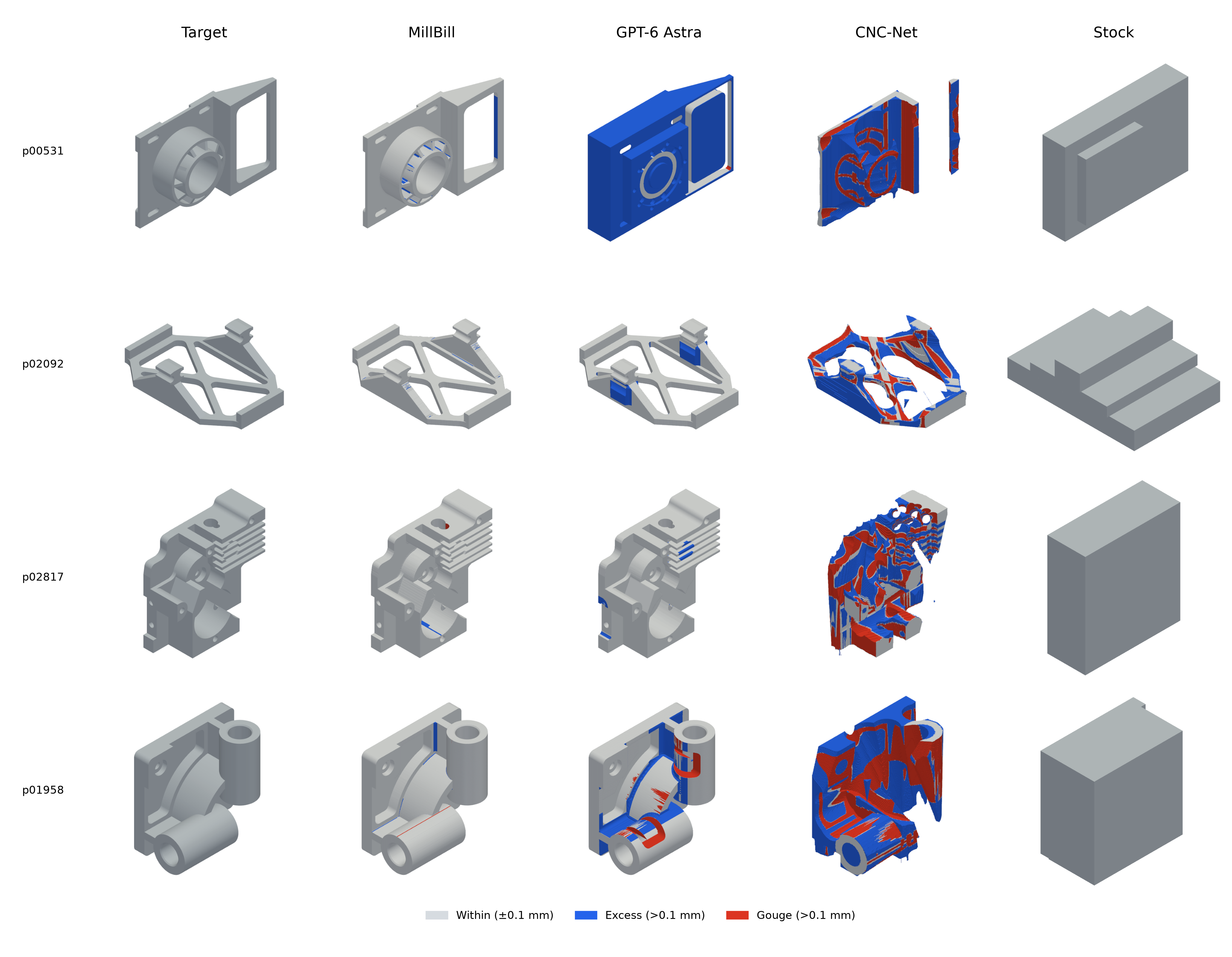}
  \caption{\textbf{Simulated results.} Four parts (rows): the target, the workpieces simulated from the programs of MillBill, the best code-writing agent (GPT-6 Astra) and CNC-Net, and the stock. Surface coloured by deviation from the design: grey within $\tau = 0.1$\,mm, blue excess material, red gouge.}
  \label{fig:qualitative}
\end{figure*}

\Cref{fig:qualitative} shows simulated parts coloured by signed deviation from the design. On p00531, the GPT-6 Astra agent leaves excess material over most of the part, while MillBill leaves it only in a few narrow regions; on p01958, the agent gouges the cylindrical bosses. CNC-Net both gouges and leaves excess material on all four parts.

\subsection{Ablations}
\label{sec:exp:ablation}

\Cref{tab:ablation} removes one component at a time. \emph{Reference plan} executes the standard-practice reference of \cref{sec:method:agent} without the language model, which separates the contribution of the model from that of the harness. \emph{Fixed $+z$ setup} replaces the reach-aware setups of \cref{sec:method:setups} with a single setup from $+z$.

\begin{table}[t]
  \centering
  \small
  \setlength{\tabcolsep}{4pt}
  \begin{tabular}{@{}lccc@{}}
    \toprule
    Variant & $\text{Acc}_\tau$ $\uparrow$ & Gouge $\downarrow$ & Excess $\downarrow$ \\
    \midrule
    MillBill (full) & \textbf{97.4} & \textbf{0.54} & \textbf{2.1} \\
    \midrule
    Reference plan (no LLM) & 97.0 & 0.77 & 2.2 \\
    Fixed $+z$ setup & 66.4 & 0.58 & 33.1 \\
    \bottomrule
  \end{tabular}
  \caption{\textbf{Ablations} on the 100-part benchmark at $\tau = 0.1$\,mm. $\text{Acc}_\tau$, gouge and excess: percentage of the target surface area of the final workpiece, averaged over the scored parts. All variants score all 100 parts.}
  \label{tab:ablation}
\end{table}

Executed without the language model (including knowledge retrieval) but with the harness corrections and repair rounds, the reference plan already leaves 97.0\% of the surface within tolerance; the model raises this to 97.4\% (paired difference 0.35\,pp, 95\% bootstrap CI [0.09, 0.68]; better on 43 parts, worse on 23) and reduces the gouged area from 0.77\% to 0.54\%. This reduction comes from the ten single-setup parts: on the 90 multi-setup parts the gouged area is 0.46\% with the model and 0.45\% without it. \Cref{sec:suppl:results} walks through one such part.

\subsection{Analysis}
\label{sec:exp:analysis}

\begin{figure}[t]
  \centering
  \begin{tabular}{@{}c@{\hspace{0.02\linewidth}}c@{}}
    \includegraphics[width=0.42\linewidth]{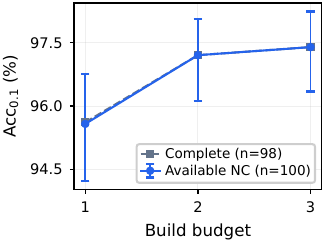} &
    \includegraphics[width=0.42\linewidth]{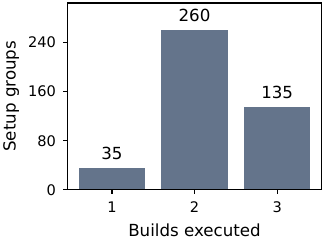} \\
    \small (a) & \small (b)
  \end{tabular}
  \caption{\textbf{Refinement.} (a)~$\text{Acc}_{0.1}$ vs.\ build budget, replayed from the saved runs (best build kept), 95\% bootstrap CIs; available: all programs; complete: the 98 parts with all setups at every budget. (b)~Builds executed per setup.}
  \label{fig:rounds}
\end{figure}

\paragraph{Refinement rounds.}
With a budget of one build, MillBill leaves 95.6\% of the surface within tolerance; a second build raises this to 97.2\% and a third to 97.4\% (\cref{fig:rounds}a); \cref{fig:rounds}b shows how many builds the 430 setups use.

\paragraph{Sensitivity to the tolerance.}
We rescore the same programs at $\tau \in \{0.05, 0.2, 0.5\}$\,mm (\cref{tab:suppl:tolerance}). MillBill stays ahead at 0.2 and 0.5\,mm (98.0\% and 98.1\%, against 95.3\% and 95.6\% for the GPT-6 Astra agent), while at 0.05\,mm the GPT-6 Astra agent is slightly ahead (90.2\% against 89.3\%).

\paragraph{Failure cases.}
MillBill's residual errors are unremoved material and local gouges (\cref{fig:suppl:failures}), up to 35.5\% excess on p05451. feature recognition fell back from AAGNet to rule-based recognition on five parts and FreeCAD timed out twice, yet every part got a program.

\section{Limitations}
\label{sec:limitations}

MillBill is limited to indexed three-axis machining from the six principal directions; parts that need inclined setups or simultaneous five-axis motion are out of scope. Workholding and machine travel are not modelled: setups are chosen by tool accessibility, not by whether the part can be clamped, the toolpaths are not checked for collisions with fixtures or the tool holder (\cref{eq:reach} is only a necessary condition for holder clearance), and parts exceeding the machine travel are not rejected. The geometric simulation does not account for tool deflection, vibration or wear. As a result, a toolpath that meets tolerance requirements in simulation may fail to achieve them during machining. Taps and thread mills are also not simulated, meaning that thread geometry is planned but not verified. The knowledge base holds representative, not vendor-specific, cutting data, and AAGNet is trained on synthetic parts, so recognition errors on real parts can propagate into the plan, though the simulation catches their geometric effects. We evaluate on 100 parts from one public library, one run per method and part.

\section{Conclusion}
\label{sec:conclusion}

We introduced MillBill, an agent that turns a B-rep part into G-code by separating machining decisions from geometric execution. A locally hosted language model refines an initial plan of operations, tools and cutting parameters. A geometric harness recognises features, selects setups by tool accessibility, generates every toolpath, simulates the cut and reports deviations per feature and operation, so the model never computes coordinates. With a 27B open model, MillBill achieves 97.4\% of the target surface within 0.1\,mm on 100 parts, against 94.3\% for the best code-writing frontier agent; the initial plan without the model reaches 97.0\%. The same simulator can score any planner's G-code. Future work includes five-axis setups, workholding and verification of threads and surface finish.

{
    \small
    \bibliographystyle{ieeenat_fullname}
    \bibliography{main}

\begin{thebibliography}{57}
\providecommand{\natexlab}[1]{#1}
\providecommand{\url}[1]{\texttt{#1}}
\expandafter\ifx\csname urlstyle\endcsname\relax
  \providecommand{\doi}[1]{doi: #1}\else
  \providecommand{\doi}{doi: \begingroup \urlstyle{rm}\Url}\fi

\bibitem[fre()]{freecad}
{FreeCAD}: Open-source parametric {3D} modeler.
\newblock \url{https://www.freecad.org}.
\newblock CAM Workbench.

\bibitem[Abdel-Malek et~al.(2006)Abdel-Malek, Yang, Blackmore, and
  Joy]{abdelmalek2006swept}
Karim Abdel-Malek, Jingzhou Yang, Denis Blackmore, and Ken Joy.
\newblock Swept volumes: Foundation, perspectives, and applications.
\newblock \emph{International Journal of Shape Modeling}, 12\penalty0
  (1):\penalty0 87--127, 2006.

\bibitem[Abdelaal et~al.(2025)Abdelaal, Lokadjaja, and Engert]{gllm}
Mohamed Abdelaal, Samuel Lokadjaja, and Gilbert Engert.
\newblock {GLLM}: Self-corrective {G-Code} generation using large language
  models with user feedback.
\newblock \emph{arXiv preprint arXiv:2501.17584}, 2025.

\bibitem[Al-wswasi et~al.(2018)Al-wswasi, Ivanov, and
  Makatsoris]{alwswasi2018acapp}
Mazin Al-wswasi, Atanas Ivanov, and Harris Makatsoris.
\newblock A survey on smart automated computer-aided process planning ({ACAPP})
  techniques.
\newblock \emph{The International Journal of Advanced Manufacturing
  Technology}, 97\penalty0 (1-4):\penalty0 809--832, 2018.

\bibitem[Alrashedy et~al.(2025)Alrashedy, Tambwekar, Zaidi, Langwasser, Xu, and
  Gombolay]{cadcodeverify}
Kamel Alrashedy, Pradyumna Tambwekar, Zulfiqar~Haider Zaidi, Megan Langwasser,
  Wei Xu, and Matthew Gombolay.
\newblock Generating {CAD} code with vision-language models for {3D} designs.
\newblock In \emph{International Conference on Learning Representations
  (ICLR)}, 2025.

\bibitem[Anderson(1978)]{anderson1978zmap}
R.~O. Anderson.
\newblock Detecting and eliminating collisions in {NC} machining.
\newblock \emph{Computer-Aided Design}, 10\penalty0 (4):\penalty0 231--237,
  1978.

\bibitem[{Anthropic}(2026)]{opus55}
{Anthropic}.
\newblock Introducing {Claude Opus 5.5}.
\newblock \url{https://www.anthropic.com/news/claude-opus-5-5}, 2026.

\bibitem[Azab et~al.(2024)Azab, Osman, and Baki]{cappgpt}
Ahmed Azab, Hany Osman, and Fazle Baki.
\newblock {CAPP-GPT}: A computer-aided process planning-generative pretrained
  transformer framework for smart manufacturing.
\newblock \emph{Manufacturing Letters}, 41:\penalty0 51--62, 2024.

\bibitem[Barkley et~al.(2026)Barkley, Loghmani, and Barati~Farimani]{cadsmith}
Jesse Barkley, Rumi Loghmani, and Amir Barati~Farimani.
\newblock {CADSmith}: Multi-agent {CAD} generation with programmatic geometric
  validation.
\newblock \emph{arXiv preprint arXiv:2603.26512}, 2026.

\bibitem[Behandish et~al.(2018)Behandish, Nelaturi, and
  de~Kleer]{behandish2018hybrid}
Morad Behandish, Saigopal Nelaturi, and Johan de Kleer.
\newblock Automated process planning for hybrid manufacturing.
\newblock \emph{Computer-Aided Design}, 102:\penalty0 115--127, 2018.

\bibitem[Cao et~al.(2020)Cao, Robinson, Hua, Boussuge, Colligan, and
  Pan]{mfcad}
Weijuan Cao, Trevor Robinson, Yang Hua, Flavien Boussuge, Andrew~R. Colligan,
  and Wanbin Pan.
\newblock Graph representation of {3D} {CAD} models for machining feature
  recognition with deep learning.
\newblock In \emph{ASME International Design Engineering Technical Conferences
  and Computers and Information in Engineering Conference (IDETC/CIE)}, page
  V11AT11A003, 2020.

\bibitem[Colligan et~al.(2022)Colligan, Robinson, Nolan, Hua, and
  Cao]{hierarchicalcadnet}
Andrew~R. Colligan, Trevor~T. Robinson, Declan~C. Nolan, Yang Hua, and Weijuan
  Cao.
\newblock Hierarchical {CADNet}: Learning from {B-Reps} for machining feature
  recognition.
\newblock \emph{Computer-Aided Design}, 147:\penalty0 103226, 2022.

\bibitem[Dai et~al.(2025)Dai, Huang, Bai, Guo, Gan, Yang, and Shi]{brepformer}
Yongkang Dai, Xiaoshui Huang, Yunpeng Bai, Hao Guo, Hongping Gan, Ling Yang,
  and Yilei Shi.
\newblock {BRepFormer}: Transformer-based {B-rep} geometric feature
  recognition.
\newblock In \emph{ACM International Conference on Multimedia Retrieval
  (ICMR)}, pages 155--163, 2025.

\bibitem[Doi and Koide(1991)]{doi1991tetrahedral}
Akio Doi and Akio Koide.
\newblock An efficient method of triangulating equi-valued surfaces by using
  tetrahedral cells.
\newblock \emph{IEICE Transactions on Information}, E74-D\penalty0
  (1):\penalty0 214--224, 1991.

\bibitem[Elistratov et~al.(2026)Elistratov, Barannikov, Ivanov, Khrulkov,
  Konushin, Kuznetsov, and Zhemchuzhnikov]{cadevolve}
Maksim Elistratov, Marina Barannikov, Gregory Ivanov, Valentin Khrulkov, Anton
  Konushin, Andrey Kuznetsov, and Dmitrii Zhemchuzhnikov.
\newblock {CADEvolve}: Creating realistic {CAD} via program evolution.
\newblock \emph{arXiv preprint arXiv:2602.16317}, 2026.

\bibitem[Fuchs et~al.(2026)Fuchs, Kacan, Haag, and Lohse]{featurefox}
Bertram Fuchs, Altay Kacan, Aaron Haag, and Oliver Lohse.
\newblock {FeatureFox}: Sample-efficient panoptic graph segmentation for
  machining feature recognition in {B-Rep} {3D-CAD} models.
\newblock In \emph{International Joint Conference on Neural Networks (IJCNN)},
  pages 1--6, 2026.

\bibitem[Gong et~al.(2026)Gong, Wu, Liu, and Tu]{toolcad}
Yifei Gong, Xing Wu, Wenda Liu, and Kang Tu.
\newblock {ToolCAD}: Exploring tool-using large language models in
  text-to-{CAD} generation with reinforcement learning.
\newblock In \emph{Annual Meeting of the Association for Computational
  Linguistics (ACL)}, 2026.

\bibitem[Hashem~Pour et~al.(2025)Hashem~Pour, Mahjourian, and
  Nguyen]{hashempour2025vlmgcode}
Yasaman Hashem~Pour, Nazanin Mahjourian, and Vinh Nguyen.
\newblock Few-shot {VLM}-based {G-Code} and {HMI} verification in {CNC}
  machining.
\newblock \emph{arXiv preprint arXiv:2512.11296}, 2025.

\bibitem[Hoang et~al.(2026)Hoang, Gorsich, Castanier, and Imani]{arkness}
Danny Hoang, David Gorsich, Matthew~P. Castanier, and Farhad Imani.
\newblock Knowledge graph fusion with large language models for accurate,
  explainable manufacturing process planning.
\newblock \emph{International Journal of Production Economics}, 300:\penalty0
  110091, 2026.

\bibitem[Hou and Frank(2017)]{hou2017gvm}
Guangyu Hou and Matthew~C. Frank.
\newblock Computing the global visibility map using slice geometry for setup
  planning.
\newblock \emph{Journal of Manufacturing Science and Engineering}, 139\penalty0
  (8):\penalty0 081006, 2017.

\bibitem[Hu et~al.(2026)Hu, Ai, Wen, Li, Zou, Li, Deng, Cai, Zhou, Cai, Fu,
  Yang, Zhang, Shi, and Yang]{itercad}
Tao Hu, Jiaxin Ai, Licheng Wen, Xueheng Li, Shu Zou, Siqi Li, Nianchen Deng,
  Xinyu Cai, Hongbin Zhou, Pinlong Cai, Daocheng Fu, Yu Yang, Hairong Zhang,
  Botian Shi, and Xuemeng Yang.
\newblock {IterCAD}: An iterative multimodal agent for visually-grounded {CAD}
  generation and editing.
\newblock \emph{arXiv preprint arXiv:2606.13368}, 2026.

\bibitem[Inui et~al.(2018)Inui, Nagano, and Umezu]{inui2018accessibility}
Masatomo Inui, Shinji Nagano, and Nobuyuki Umezu.
\newblock Fast computation of accessibility cones for assisting 3+2 axis
  milling.
\newblock \emph{Computer-Aided Design and Applications}, 15\penalty0
  (5):\penalty0 667--676, 2018.

\bibitem[Jayaraman et~al.(2021)Jayaraman, Sanghi, Lambourne, Willis, Davies,
  Shayani, and Morris]{uvnet}
Pradeep~Kumar Jayaraman, Aditya Sanghi, Joseph~G. Lambourne, Karl D.~D. Willis,
  Thomas Davies, Hooman Shayani, and Nigel Morris.
\newblock {UV-Net}: Learning from boundary representations.
\newblock In \emph{IEEE/CVF Conference on Computer Vision and Pattern
  Recognition (CVPR)}, pages 11698--11707, 2021.

\bibitem[Jignasu et~al.(2023)Jignasu, Marshall, Ganapathysubramanian, Balu,
  Hegde, and Krishnamurthy]{jignasu2023gcode}
Anushrut Jignasu, Kelly Marshall, Baskar Ganapathysubramanian, Aditya Balu,
  Chinmay Hegde, and Adarsh Krishnamurthy.
\newblock Towards foundational {AI} models for additive manufacturing: Language
  models for {G-Code} debugging, manipulation, and comprehension.
\newblock \emph{arXiv preprint arXiv:2309.02465}, 2023.

\bibitem[Joy and Feng(2017)]{joy2017voxel}
Jimin Joy and Hsi-Yung Feng.
\newblock Efficient milling part geometry computation via three-step update of
  frame-sliced voxel representation workpiece model.
\newblock \emph{The International Journal of Advanced Manufacturing
  Technology}, 92\penalty0 (5-8):\penalty0 2365--2378, 2017.

\bibitem[Kabisov et~al.(2026{\natexlab{a}})Kabisov, Kirichuk, Volkov, Savrasov,
  Barannikov, Konushin, Kuznetsov, and Zhemchuzhnikov]{cadreasoner}
Soslan Kabisov, Vsevolod Kirichuk, Andrey Volkov, Gennadii Savrasov, Marina
  Barannikov, Anton Konushin, Andrey Kuznetsov, and Dmitrii Zhemchuzhnikov.
\newblock {CADReasoner}: Iterative program editing for {CAD} reverse
  engineering.
\newblock In \emph{IEEE/CVF Conference on Computer Vision and Pattern
  Recognition (CVPR), Findings}, 2026{\natexlab{a}}.

\bibitem[Kabisov et~al.(2026{\natexlab{b}})Kabisov, Savrasov, Elistratov,
  Rodriguez, Ignatiev, Gavrilov, Uzdenov, Boyko, Pasechnik, Konushin,
  Kuznetsov, and Zhemchuzhnikov]{cadena}
Soslan Kabisov, Gennadiy Savrasov, Maksim Elistratov, Antonio Rodriguez, Daniil
  Ignatiev, Nikita Gavrilov, Rustam Uzdenov, Alexey~I. Boyko, Igor Pasechnik,
  Anton Konushin, Andrey Kuznetsov, and Dmitrii Zhemchuzhnikov.
\newblock {CADENA}: Stepwise {CAD} reverse engineering.
\newblock \emph{arXiv preprint arXiv:2608.00799}, 2026{\natexlab{b}}.

\bibitem[Khan et~al.(2025)Khan, Chen, Ng, Feng, Tan, and Moon]{khan2025vlmfr}
Muhammad~Tayyab Khan, Lequn Chen, Ye~Han Ng, Wenhe Feng, Nicholas Yew~Jin Tan,
  and Seung~Ki Moon.
\newblock Leveraging vision-language models for manufacturing feature
  recognition in computer-aided designs.
\newblock \emph{Journal of Computing and Information Science in Engineering},
  25\penalty0 (10):\penalty0 104501, 2025.

\bibitem[Khan et~al.(2026)Khan, Chen, Feng, and Moon]{designtoplan}
Muhammad~Tayyab Khan, Lequn Chen, Wenhe Feng, and Seung~Ki Moon.
\newblock Design-to-plan: A large language model-based multi-agent framework
  for manufacturing process planning from {3D} {CAD} models and {2D}
  engineering drawings.
\newblock \emph{arXiv preprint arXiv:2608.24039}, 2026.

\bibitem[Kienzle(1952)]{kienzle1952}
O. Kienzle.
\newblock Die {B}estimmung von {K}r{\"a}ften und {L}eistungen an spanenden
  {W}erkzeugen und {W}erkzeugmaschinen.
\newblock \emph{VDI-Z}, 94\penalty0 (11/12):\penalty0 299--305, 1952.

\bibitem[Kolodiazhnyi et~al.(2025)Kolodiazhnyi, Tarasov, Zhemchuzhnikov,
  Nikulin, Zisman, Vorontsova, Konushin, Kurenkov, and Rukhovich]{cadrille}
Maksim Kolodiazhnyi, Denis Tarasov, Dmitrii Zhemchuzhnikov, Alexander Nikulin,
  Ilya Zisman, Anna Vorontsova, Anton Konushin, Vladislav Kurenkov, and Danila
  Rukhovich.
\newblock cadrille: Multi-modal {CAD} reconstruction with reinforcement
  learning.
\newblock \emph{arXiv preprint arXiv:2505.22914}, 2025.

\bibitem[Lambourne et~al.(2021)Lambourne, Willis, Jayaraman, Sanghi, Meltzer,
  and Shayani]{brepnet}
Joseph~G. Lambourne, Karl D.~D. Willis, Pradeep~Kumar Jayaraman, Aditya Sanghi,
  Peter Meltzer, and Hooman Shayani.
\newblock {BRepNet}: A topological message passing system for solid models.
\newblock In \emph{IEEE/CVF Conference on Computer Vision and Pattern
  Recognition (CVPR)}, pages 12768--12777, 2021.

\bibitem[Lee(2026)]{lee2026gcode}
Yeonseok Lee.
\newblock Correct-by-construction {G-Code} generation: A neuro-symbolic
  approach via separation logic.
\newblock \emph{arXiv preprint arXiv:2605.10568}, 2026.

\bibitem[Li et~al.(2024)Li, Zhao, Jiang, Pan, Liu, Wu, Shu, Tian, Yang, Xu,
  Lyu, Blenk, Pence, Rupram, Banu, Liu, Wang, Song, Zhai, Song, Zhu, Li, Wang,
  and Liu]{llm4manufacturing}
Yiwei Li, Huaqin Zhao, Hanqi Jiang, Yi Pan, Zhengliang Liu, Zihao Wu, Peng Shu,
  Jie Tian, Tianze Yang, Shaochen Xu, Yanjun Lyu, Parker Blenk, Jacob Pence,
  Jason Rupram, Eliza Banu, Ninghao Liu, Linbing Wang, Wenzhan Song, Xiaoming
  Zhai, Kenan Song, Dajiang Zhu, Beiwen Li, Xianqiao Wang, and Tianming Liu.
\newblock Large language models for manufacturing.
\newblock \emph{arXiv preprint arXiv:2410.21418}, 2024.

\bibitem[Madaan et~al.(2023)Madaan, Tandon, Gupta, Hallinan, Gao, Wiegreffe,
  Alon, Dziri, Prabhumoye, Yang, Gupta, Majumder, Hermann, Welleck,
  Yazdanbakhsh, and Clark]{selfrefine}
Aman Madaan, Niket Tandon, Prakhar Gupta, Skyler Hallinan, Luyu Gao, Sarah
  Wiegreffe, Uri Alon, Nouha Dziri, Shrimai Prabhumoye, Yiming Yang, Shashank
  Gupta, Bodhisattwa~Prasad Majumder, Katherine Hermann, Sean Welleck, Amir
  Yazdanbakhsh, and Peter Clark.
\newblock Self-refine: Iterative refinement with self-feedback.
\newblock In \emph{Advances in Neural Information Processing Systems
  (NeurIPS)}, 2023.

\bibitem[Mallis et~al.(2025)Mallis, Karadeniz, Cavada, Rukhovich,
  Foteinopoulou, Cherenkova, Kacem, and Aouada]{cadassistant}
Dimitrios Mallis, Ahmet~Serdar Karadeniz, Sebastian Cavada, Danila Rukhovich,
  Niki Foteinopoulou, Kseniya Cherenkova, Anis Kacem, and Djamila Aouada.
\newblock {CAD-Assistant}: Tool-augmented {VLLMs} as generic {CAD} task
  solvers.
\newblock In \emph{IEEE/CVF International Conference on Computer Vision
  (ICCV)}, 2025.

\bibitem[Nelaturi et~al.(2015)Nelaturi, Burton, Fritz, and
  Kurtoglu]{nelaturi2015spatial}
Saigopal Nelaturi, Gregory Burton, Christian Fritz, and Tolga Kurtoglu.
\newblock Automatic spatial planning for machining operations.
\newblock In \emph{IEEE International Conference on Automation Science and
  Engineering (CASE)}, pages 677--682, 2015.

\bibitem[{OpenAI}(2026)]{gpt6astra}
{OpenAI}.
\newblock {GPT-6 Astra} system card.
\newblock \url{https://deploymentsafety.openai.com/gpt-6-astra}, 2026.

\bibitem[Park(2004)]{park2004zlevel}
Sang~C. Park.
\newblock Sculptured surface machining using triangular mesh slicing.
\newblock \emph{Computer-Aided Design}, 36\penalty0 (3):\penalty0 279--288,
  2004.

\bibitem[{Qwen Team}(2026)]{qwen38}
{Qwen Team}.
\newblock {Qwen3.8-Max}: A new bar for coding and cowork.
\newblock \url{https://qwen.ai/blog?id=qwen3.8}, 2026.

\bibitem[Savrasov et~al.(2026)Savrasov, Elistratov, Gavrilov, Garifullin,
  Pavlov, Kabisov, Frolov, Konushin, Kuznetsov, and Zhemchuzhnikov]{cadfather}
Gennadiy Savrasov, Maksim Elistratov, Nikita Gavrilov, Albert Garifullin, Oleg
  Pavlov, Soslan Kabisov, Vladimir Frolov, Anton Konushin, Andrey Kuznetsov,
  and Dmitrii Zhemchuzhnikov.
\newblock {CADFather}: Autonomous {CAD} reconstruction through coordinated tool
  use.
\newblock \emph{arXiv preprint arXiv:2610.09127}, 2026.

\bibitem[Schick et~al.(2023)Schick, Dwivedi-Yu, Dess{\`i}, Raileanu, Lomeli,
  Zettlemoyer, Cancedda, and Scialom]{toolformer}
Timo Schick, Jane Dwivedi-Yu, Roberto Dess{\`i}, Roberta Raileanu, Maria
  Lomeli, Luke Zettlemoyer, Nicola Cancedda, and Thomas Scialom.
\newblock Toolformer: Language models can teach themselves to use tools.
\newblock \emph{arXiv preprint arXiv:2302.04761}, 2023.

\bibitem[Schn{\"o}s et~al.(2021)Schn{\"o}s, Hartmann, Obst, and
  Glashagen]{schnoes2021gpuvoxel}
Florian Schn{\"o}s, Dirk Hartmann, Birgit Obst, and Glenn Glashagen.
\newblock {GPU} accelerated voxel-based machining simulation.
\newblock \emph{The International Journal of Advanced Manufacturing
  Technology}, 115\penalty0 (1-2):\penalty0 275--289, 2021.

\bibitem[Shinn et~al.(2023)Shinn, Cassano, Berman, Gopinath, Narasimhan, and
  Yao]{reflexion}
Noah Shinn, Federico Cassano, Edward Berman, Ashwin Gopinath, Karthik
  Narasimhan, and Shunyu Yao.
\newblock Reflexion: Language agents with verbal reinforcement learning.
\newblock \emph{arXiv preprint arXiv:2303.11366}, 2023.

\bibitem[{\v{S}}ket et~al.(2025){\v{S}}ket, Poto{\v{c}}nik, Brezocnik, Ficko,
  and Klan{\v{c}}nik]{sket2025gcode}
K. {\v{S}}ket, D. Poto{\v{c}}nik, M. Brezocnik, M. Ficko, and S.
  Klan{\v{c}}nik.
\newblock Large language models for {G-code} generation in {CNC} machining: A
  comparison of {ChatGPT-3.5} and {ChatGPT-4o}.
\newblock \emph{Advances in Production Engineering \& Management}, 20\penalty0
  (2):\penalty0 224--238, 2025.

\bibitem[Van~Hook(1986)]{vanhook1986dexel}
Tim Van~Hook.
\newblock Real-time shaded {NC} milling display.
\newblock \emph{ACM SIGGRAPH Computer Graphics}, 20\penalty0 (4):\penalty0
  15--20, 1986.

\bibitem[Woo(1994)]{woo1994visibility}
Tony~C. Woo.
\newblock Visibility maps and spherical algorithms.
\newblock \emph{Computer-Aided Design}, 26\penalty0 (1):\penalty0 6--16, 1994.

\bibitem[Wu et~al.(2024)Wu, Lei, Peng, and Gao]{aagnet}
Hongjin Wu, Ruoshan Lei, Yibing Peng, and Liang Gao.
\newblock {AAGNet}: A graph neural network towards multi-task machining feature
  recognition.
\newblock \emph{Robotics and Computer-Integrated Manufacturing}, 86:\penalty0
  102661, 2024.

\bibitem[Xu et~al.(2011)Xu, Wang, and Newman]{xu2011capp}
Xun Xu, Lihui Wang, and Stephen~T. Newman.
\newblock Computer-aided process planning -- a critical review of recent
  developments and future trends.
\newblock \emph{International Journal of Computer Integrated Manufacturing},
  24\penalty0 (1):\penalty0 1--31, 2011.

\bibitem[Yao et~al.(2026)Yao, Wu, Zheng, Xing, and Fu]{brepmae}
Can Yao, Kang Wu, Zuheng Zheng, Siyuan Xing, and Xiao-Ming Fu.
\newblock {BRepMAE}: Self-supervised masked {BRep} autoencoders for machining
  feature recognition.
\newblock \emph{arXiv preprint arXiv:2602.22701}, 2026.

\bibitem[Yao et~al.(2023)Yao, Zhao, Yu, Du, Shafran, Narasimhan, and
  Cao]{react}
Shunyu Yao, Jeffrey Zhao, Dian Yu, Nan Du, Izhak Shafran, Karthik Narasimhan,
  and Yuan Cao.
\newblock {ReAct}: Synergizing reasoning and acting in language models.
\newblock In \emph{International Conference on Learning Representations
  (ICLR)}, 2023.

\bibitem[Yao et~al.(2003)Yao, Gupta, and Nau]{yao2003cutters}
Zhiyang Yao, Satyandra~K. Gupta, and Dana~S. Nau.
\newblock Algorithms for selecting cutters in multi-part milling problems.
\newblock \emph{Computer-Aided Design}, 35\penalty0 (9):\penalty0 825--839,
  2003.

\bibitem[Yavartanoo et~al.(2024)Yavartanoo, Hong, Neshatavar, and Lee]{cncnet}
Mohsen Yavartanoo, Sangmin Hong, Reyhaneh Neshatavar, and Kyoung~Mu Lee.
\newblock {CNC-Net}: Self-supervised learning for {CNC} machining operations.
\newblock In \emph{Proceedings of the IEEE/CVF Conference on Computer Vision
  and Pattern Recognition (CVPR)}, pages 9816--9825, 2024.

\bibitem[Zhang et~al.(2024)Zhang, Guan, Jiang, Wang, and Tan]{brepmfr}
Shuming Zhang, Zhidong Guan, Hao Jiang, Xiaodong Wang, and Pingan Tan.
\newblock {BrepMFR}: Enhancing machining feature recognition in {B-rep} models
  through deep learning and domain adaptation.
\newblock \emph{Computer Aided Geometric Design}, 111:\penalty0 102318, 2024.

\bibitem[Zhang and Leu(2009)]{zhang2009tridexel}
Weihan Zhang and Ming~C. Leu.
\newblock Surface reconstruction using dexel data from three sets of orthogonal
  rays.
\newblock \emph{Journal of Computing and Information Science in Engineering},
  9\penalty0 (1):\penalty0 011008, 2009.

\bibitem[Zhang et~al.(2018)Zhang, Jaiswal, and Rai]{featurenet}
Zhibo Zhang, Prakhar Jaiswal, and Rahul Rai.
\newblock {FeatureNet}: Machining feature recognition based on {3D} convolution
  neural network.
\newblock \emph{Computer-Aided Design}, 101:\penalty0 12--22, 2018.

\bibitem[Zhou et~al.(2026)Zhou, Lin, Feng, and Zheng]{cncgen}
Xiaolei Zhou, Boyi Lin, Yuchao Feng, and Jianwei Zheng.
\newblock {CNCGEN}: A dataset and framework for machining process planning and
  toolpath generation from {B-rep} models.
\newblock \emph{arXiv preprint arXiv:2609.39738}, 2026.

\end{thebibliography}
}

\appendix
\setcounter{section}{0}\setcounter{table}{0}\setcounter{figure}{0}\setcounter{equation}{0}
\renewcommand{\thesection}{S\arabic{section}}\renewcommand{\theHsection}{S\arabic{section}}
\renewcommand{\thetable}{S\arabic{table}}\renewcommand{\theHtable}{S\arabic{table}}
\renewcommand{\thefigure}{S\arabic{figure}}\renewcommand{\theHfigure}{S\arabic{figure}}
\renewcommand{\theequation}{S\arabic{equation}}\renewcommand{\theHequation}{S\arabic{equation}}
\clearpage
\maketitlesupplementary

\section{Implementation details}
\label{sec:suppl:impl}

\paragraph{Language model and agents.}
All agents use Qwen3.8-27B~\cite{qwen38} served locally with vLLM through an OpenAI-compatible interface, in FP8 on four NVIDIA H100 80\,GB GPUs with tensor parallelism of four, with a 250{,}000-token context (54.6 calls per part on average on the benchmark), at temperature 0 with thinking disabled. Structured answers are requested in JSON mode and retried up to twice if they do not parse. Each ReAct agent~\cite{react} has a fixed iteration budget (\cref{tab:suppl:params}); on the last iteration tools are withheld so that the model must answer. Identical repeated tool calls return the cached result with a warning, and a turn consisting only of repeated calls forces the final answer. Tool results are truncated to 3{,}500 characters in the conversation. Knowledge retrieval runs three sub-agents concurrently (text and knowledge graph; SQL, tables and decision tree; materials database), whose outputs are merged deterministically and summarised by a coordinator call.

\paragraph{Knowledge base.}
The knowledge base contains 17 workpiece materials with ISO~513 groups and Kienzle constants, 26 machining rules and 37 thread specifications (ISO~261, ASME~B1.1) in SQL tables, 55 feature--material rows, 23 sections of machining guidelines retrieved by BM25, and a knowledge graph of 103 triples over processes, features and materials. The tool library has 75 cutters: 30 flat, 7 ball and 5 bull-nose end mills, 5 face mills, 5 twist drills, 4 spot drills, 6 countersinks, 3 chamfer mills, 3 taps, 3 thread mills, 2 reamers and 2 boring bars.

\paragraph{Feature recognition.}
AAGNet~\cite{aagnet} runs in its own environment with the authors' MFInstSeg weights. Instance affinities and bottom-face probabilities are thresholded at 0.5. A predicted face is matched to a CAD-kernel face if their centroids are within $0.05\,\text{mm} + 10^{-3}$ times the part scale and their areas within 1\%. \Cref{tab:suppl:aagnet} lists how MFInstSeg classes map to the feature types used in planning.

\begin{table}[h]
  \centering
  \small
  \begin{tabular}{@{}lp{0.72\linewidth}@{}}
    \toprule
    Type & MFInstSeg classes \\
    \midrule
    hole    & through hole, blind hole \\
    pocket  & triangular, rectangular and 6-sided pockets; circular-end pocket; triangular, rectangular and 6-sided passages; O-ring \\
    slot    & triangular, rectangular and circular through slots; rectangular blind slot; v and h circular-end blind slots \\
    step    & rectangular, 2-sided and slanted through steps; triangular, circular and rectangular blind steps \\
    chamfer & chamfer \\
    fillet  & round \\
    \bottomrule
  \end{tabular}
  \caption{Mapping of the 24 MFInstSeg classes to feature types; the stock class is ignored. Later geometric checks may refine the type.}
  \label{tab:suppl:aagnet}
\end{table}

\paragraph{Annotation fusion.}
In \cref{eq:fusion} the weights are $(w_{\text{type}}, w_{\text{dim}}, w_{\text{loc}}) = (0.35, 0.5, 0.15)$, or $(0.6, 0.1, 0.3)$ when the annotation carries no dimension or thread. Pairs scoring above 0.2 are candidates; a match is accepted without the model if its score is at least 0.75 and it is unique, its count matches, or it is a thread or hole annotation. Model resolutions with confidence below 0.45 are flagged for review.

\paragraph{Reachability.}
The holder diameter is 40\,mm and the clamped length of every cutter is 25\,mm. The reachability test of \cref{eq:reach} uses the flat, ball and bull-nose end mills of the library and runs on voxels of at least 0.3\,mm (at most $9\times10^6$ voxels) with part tops max-pooled on a 2\,mm lattice, which makes it conservative by at most one coarse cell. A feature is assigned to a setup if at least 80\% of its volume is reachable from it and at least 50\% is not yet removed. A further setup is added while its new reachable volume exceeds $\max(30\,\text{mm}^3, 0.3\%$ of the part volume$)$; $+z$ is preferred as the primary setup if it reaches at least 90\% of the best direction's volume. The reach test checks holder clearance and tool length only; whether a cutter fits a feature is decided later by tool selection.

\paragraph{Cutting data.}
Spindle speed is $n = 1000\,v_c / (\pi D)$, capped at the machine maximum, with the cutting speed $v_c$ taken from the material and tool material. For end mills, the feed per tooth is $f_z = k\,D$ (clamped to $[0.005, 0.25]$\,mm and reduced for HSS, ball-nose and finishing cutters); at radial engagement $a_e < D/2$ it is increased by the chip-thinning factor $D / \bigl(2\sqrt{a_e (D - a_e)}\bigr)$, capped at 2.5. Axial and radial depths of cut come from engagement tables per strategy (adaptive, roughing, slotting, finishing, facing) and ISO group. The specific cutting force is $k_c = k_{c1.1}\, h_m^{-m_c}$ with mean chip thickness $h_m$, and the cutting power $P_c = a_p a_e v_f k_c / (60\cdot10^6)$\,kW; the axial depth is reduced if $P_c$ exceeds the spindle power times its efficiency. Long tools (stick-out over five diameters) get reduced speed, feed and depth. Drills peck when the depth exceeds three (HSS) or five (carbide) diameters, and tapping uses a feed equal to pitch times spindle speed.

\paragraph{Toolpaths and gouge check.}
Harness toolpaths are computed on a height map with cell $\max(0.1\,\text{mm}, \ell/900)$, where $\ell$ is the largest part dimension. Roughing leaves a 0.3\,mm allowance when a finishing step follows. Finishing waterlines are spaced at $\mathrm{clamp}(D/4, 0.3, 1.5)$\,mm on a distance transform at the verification resolution. The gouge check splits moves into pieces of at most $\max(2\,\text{cell}, 0.25\,\text{mm})$, uses a margin of $\tau/4$, and lifts violating pieces in steps of 0.01\,mm. It covers the toolpaths computed by the harness without arcs; FreeCAD operations and helical paths are checked by the verifier only. Rest finishing chooses up to two cutters among ball (10, 6, 3\,mm), bull-nose (16, 10\,mm) and flat (20, 12, 6\,mm) end mills and uses a scallop height of $\tau/4$. Tapping cycles in the post-processed G-code are rewritten to rigid tapping with the feed equal to the pitch times the spindle speed, and reaming and boring cycles to feed-out cycles. The safety check flags non-rigid tapping, rapid retracts from reamed holes and rapid moves below the stock top.

\paragraph{Simulation and metric.}
The default tolerance is $\tau = 0.1$\,mm. An arc of radius $r$ and angle $\Delta\theta$ is replaced by $n = \max\bigl(2, \lceil \Delta\theta / (2\arccos(1 - \varepsilon/r)) \rceil\bigr)$ chords, so that the sagitta is at most $\varepsilon = 0.01$\,mm. Taps and thread mills are not simulated. Cutter profiles in \cref{eq:removes} are $\eta(d) = 0$ for flat end mills, $R - \sqrt{R^2 - d^2}$ for ball end mills, $0$ for $d \le R - r_c$ and $r_c - \sqrt{r_c^2 - (d - R + r_c)^2}$ otherwise for bull-nose end mills with corner radius $r_c$, and $\max(d - r_t, 0)\cot(\theta/2)$ for conical tools with tip radius $r_t$ and point angle $\theta$. For a move with planar displacement $\mathbf{u}$ and $\mathbf{w} = \mathbf{p} - \mathbf{c}_0$ in the plane, the interval with $d(t) \le R$ is $[\max((b - \sqrt{D})/a, 0), \min((b + \sqrt{D})/a, 1)]$ with $a = \mathbf{u}\cdot\mathbf{u}$, $b = \mathbf{w}\cdot\mathbf{u}$, $D = b^2 - a(\mathbf{w}\cdot\mathbf{w} - R^2)$, and is empty if $D < 0$. A point below the lower tip height at the ends of this interval is rejected immediately. Golden-section search stops when the edge height drops to $p_z$ or the bracket is shorter than $10^{-10}$, after at most 60 steps. Moves are binned in square planar cells of side $\max(2\,\text{mm}, \sqrt{A/(4\cdot10^6)})$, where $A$ is the planar area of the stock, by their footprint dilated by $R$, and the moves of a cell are visited in order of their lowest tip height; the search stops at the first move that removes the point or whose lowest tip lies above it. A box stock is tested exactly; other stock is tessellated with deflection 0.05\,mm and tested by ray parity. When the first setup of a part is verified before the later programs exist, the later setups are assumed to remove all material above the target as seen from their direction, within the part bounds plus 2\,mm; scores of the whole part use only setups that have a program.

$N = 600{,}000$ surface points are distributed over the faces in proportion to area, placed uniformly in the triangles of each face's tessellation with a fixed seed, and projected onto the exact surface; the weight of a point is the face area divided by the number of points on the face. Without the projection, a point on a chord of a curved face lies inside the surface and its probes measure the wrong distance. Probes are taken from the sample point itself, without lateral offsets. Beyond $\rho_\tau$, probes at $1.05\tau, 1.5\tau, 2\tau, 3\tau, 5\tau$ and at 1, 2, 3, 5, 10 and 20\,mm measure the size of a deviation without changing the verdict, and each is evaluated only if all previous probes show a deviation. We also report the area out of tolerance by less than 5\%, \ie, passing with the additional probe at $1.05\tau$. The sampling error of \cref{eq:metric} is about $100\sqrt{p(1-p)/N} \le 0.065$ percentage points.

\paragraph{Visualisation mesh.}
For visualisation, we reconstruct a triangle mesh of the final workpiece boundary using marching tetrahedra~\cite{doi1991tetrahedral}. We query material membership after all setups at the nodes of a spatial grid and split each cubic cell into six tetrahedra; on edges whose endpoints lie on opposite sides of the material boundary, surface vertex positions are refined by bisection with repeated simulator queries. The resulting mesh approximates the surface at a resolution limited by the grid spacing. The metric of \cref{eq:metric} is computed directly from material-membership queries to the simulator and does not depend on this triangulation.

\paragraph{Refinement.}
Builds are ranked lexicographically by the number of safety issues, the gouged area (differences below $\max(0.05$ percentage points$, 10\%)$ are treated as equal), and then the score of \cref{eq:metric}. At most two refinement rounds follow the first build. Payloads sent to the model are truncated to 9{,}000 characters and include up to eight error regions, 20 issues and 10 operation errors.

\begin{table}[t]
  \centering
  \small
  \setlength{\tabcolsep}{4pt}
  \begin{tabular}{@{}lr@{}}
    \toprule
    Parameter & Value \\
    \midrule
    Model & Qwen3.8-27B, FP8 \\
    Temperature / thinking & 0 / off \\
    Max.\ output tokens & 8192 \\
    ReAct iter.: retrieval sub-agents & 5 / 5 / 3 \\
    ReAct iter.: sequence, tools & 5, 5 \\
    Tolerance $\tau$ & 0.1\,mm \\
    Arc sagitta $\varepsilon$ & 0.01\,mm \\
    Surface samples $N$ & 600{,}000 \\
    Builds per setup & $\le 3$ \\
    Stock allowance: side / top / bottom & 2 / 1 / 0\,mm \\
    Holder diam. / clamped length & 40 / 25\,mm \\
    Machine: max.\ spindle speed & 12{,}000\,rpm \\
    Machine: max.\ feed & 10{,}000\,mm/min \\
    Machine: power & 11\,kW \\
    \bottomrule
  \end{tabular}
  \caption{Default parameters.}
  \label{tab:suppl:params}
\end{table}

\section{Benchmark construction}
\label{sec:suppl:data}

\paragraph{Collection and licence.}
The models were obtained from the GrabCAD Community Library (\url{https://grabcad.com/library}). We selected 4{,}294 models at random; each passed a single-solid check, and all geometry is in millimetres. The models serve as geometric references for research experiments. Their use is subject to the GrabCAD Terms of Use and to the conditions specified by the respective creators; no uniform open-source licence is assumed. We acknowledge the original creators and provide links to the corresponding model pages wherever applicable.

\paragraph{Stocks.}
A part is first classed as milled from a turned or a prismatic blank. A turned blank is a stepped shaft with 1\,mm of allowance on each diameter and face, with diameters rounded up to 0.5\,mm. Where milling only opens a local cavity, the shaft is replaced by a solid of revolution with no allowance: the cavity is filled out to the circumscribed circle, while threads and knurls stay as turned. A part milled from a prismatic blank is assigned the first matching kind. A long part of constant non-rectangular section (length at least 2.5 times the width) gets a profile. A thin uniform wall is cut from sheet. A flat box, with thickness at most a quarter of the width, gets a plate, or a sheet if the part is already at sheet gauge. A part that fills at least 40\% of its bounding box is cut from bar: square bar when the two sides of the cross-section differ by at most 15\%, otherwise the section is a flat bar if the EN~10058 size lists cover it, \ie, list a width of at least the required width, up to 150\,mm, and a thickness of at least the required thickness, between 5 and 60\,mm and smaller than that width. Any other section is a plate strip cut from EN~10029 plate, with the width rounded up to the next millimetre and the smallest plate thickness that covers the required thickness; a section wider than 150\,mm is a plate. Any other part is a stepped casting or a closed-die forging: a forging if its wall is at least 0.2 of the smallest box dimension and it is at most 600\,mm long, and a casting otherwise. Allowances follow the practice named in the stock record: ISO~8062-3 for castings, EN~10243-1 for forgings, and the EN size series for bar, plate and sheet. Eleven of the 100 are milled from a turned stock: four from a stepped shaft with 1\,mm of allowance and seven from a solid of revolution with none. %

\paragraph{Near-duplicates.}
The collection itself is not deduplicated. For the 100-part draw, candidates with the same number of faces and with two of the three sorted bounding-box edges equal to within 0.01\,mm are treated as copies, mirrors or one profile cut to several lengths, and only the one with the lowest identifier is kept. This removes 112 candidates.

\paragraph{Materials.}
The workpiece material is the one specified by the author of the model on GrabCAD; candidates whose material is not in the knowledge base are not drawn. Materials are AL 6061-T6 (22), C45E / AISI 1045 (21), S235JR (16), EN AW-7075 (12), AISI 304 (10), Ti-6Al-4V (10) and POM-C (9). %

\section{Frontier-model baseline}
\label{sec:suppl:baseline}

For the comparison with frontier models, each task is exported as the part and stock in STEP format, the material, notes and tolerance, the machine profile, the coordinate convention and the tool library. The prompt asks the model to act as a CNC process planner and CAM programmer for a three-axis machining centre and to return, per setup, the side it machines from, a G-code file (G21/G90/G17, one tool change per tool, an operation comment before each operation) and a JSON description of the tools and operations. The setup frame is the same for both systems: the part is rotated so that the setup direction points up, with the minimum of its bounding box at the origin. Answers from the frontier models and from MillBill are scored by the same code with \cref{eq:metric}, together with the gouged and excess areas (\cref{tab:suppl:geometry}).
The agents are built on GPT-6 Astra and Claude Opus 5.5. In their harness, the model parses the STEP files and writes the G-code itself, with Python, image viewing and file access, and receives no feedback on its programs. Each agent has a budget of \$2 and at most 80 turns per part, with medium reasoning effort and at most 32{,}000 output tokens per response; the provider-side hardware is unknown. All Opus runs and 98 Astra runs end with a submitted answer; on \texttt{p04689} the Astra agent exhausts its budget, and its saved program is scored. The Qwen3.8-27B code-writing agent runs on our hardware with no monetary cap and at most 160 calls (80, then 40 for each of two repair rounds that return feedback from our simulator) and four hours per part; it produces valid G-code for 82 parts.

\paragraph{CNC-Net.}
The published code of CNC-Net~\cite{cncnet} simulates the material removed by its predicted operations and saves the result as an OBJ mesh; it does not convert the operations to G-code and does not support a given stock, so it starts from the bounding box of the part. We score the saved mesh with the metric of \cref{sec:suppl:impl}. It produces no output geometry for \texttt{p04119}. It is trained from scratch for each part, without a pretrained checkpoint, on one H100 GPU with seed 0, on a $64^3$ grid with export at $256^3$, and at most 100{,}000 optimisation steps with early stopping (reached on \texttt{p00178}; training fails on \texttt{p04119}).

\section{Detailed comparison}
\label{sec:suppl:geometry}

\Cref{tab:suppl:geometry} extends \cref{tab:geometry-comparison} with the scores of the initial stock, volume proxies, maximum deviation depths and the boundary band of \cref{sec:suppl:impl}. On the six parts longer than the travel of a typical three-axis machining centre (\cref{sec:exp:setup}), MillBill leaves 98.8\% of the surface within tolerance, against 99.5\% for the GPT-6 Astra agent, which fails on the longest part (1{,}840\,mm) and is scored on the other five, 82.3\% for the Opus 5.5 agent and 27.3\% for CNC-Net.

\begin{table}[t]
  \centering
  \footnotesize
  \setlength{\tabcolsep}{2.5pt}
  \begin{tabular}{@{}lrrrrr@{}}
    \toprule
    Metric & MillBill & Astra & Opus & Qwen & CNC-Net \\
    \midrule
    Scored parts & 100 & 99 & 100 & 82 & 99 \\
    \midrule
    \multicolumn{6}{@{}l}{\textit{Initial stock}} \\
    Within tolerance (\%) & 9.3 & 9.3 & 9.3 & 10.6 & 41.8 \\
    Gouge area (\%) & 0.00 & 0.00 & 0.00 & 0.00 & 0.02 \\
    Excess area (\%) & 90.7 & 90.7 & 90.7 & 89.4 & 58.2 \\
    Gouge proxy (cm$^3$) & 0.00 & 0.00 & 0.00 & 0.00 & 0.01 \\
    Excess proxy (cm$^3$) & 503.28 & 477.58 & 503.56 & 413.42 & 560.23 \\
    Max.\ gouge depth (mm) & 0.00 & 0.40 & 0.40 & 0.00 & 4.65 \\
    Max.\ excess depth (mm) & 19.45 & 19.44 & 19.45 & 19.19 & 19.27 \\
    \midrule
    \multicolumn{6}{@{}l}{\textit{Final workpiece}} \\
    Within tolerance (\%) & 97.4 & 94.3 & 83.0 & 7.9 & 42.3 \\
    Gouge area (\%) & 0.54 & 1.50 & 4.76 & 38.99 & 31.11 \\
    Excess area (\%) & 2.1 & 4.2 & 12.3 & 53.5 & 26.6 \\
    Gouge proxy (cm$^3$) & 2.86 & 31.95 & 21.01 & 168.67 & 412.65 \\
    Excess proxy (cm$^3$) & 2.33 & 9.61 & 61.35 & 254.10 & 98.16 \\
    Max.\ gouge depth (mm) & 19.40 & 16.17 & 17.02 & 17.28 & 17.98 \\
    Max.\ excess depth (mm) & 9.68 & 11.01 & 12.64 & 17.24 & 18.25 \\
    \midrule
    \multicolumn{6}{@{}l}{\textit{Boundary band (factor 1.05)}} \\
    Initial within band (\%) & 10.4 & 10.3 & 10.4 & 10.8 & 42.3 \\
    Initial boundary band (\%) & 1.1 & 1.1 & 1.1 & 0.2 & 0.5 \\
    Final within band (\%) & 97.6 & 94.4 & 83.5 & 7.9 & 42.6 \\
    Final boundary band (\%) & 0.2 & 0.1 & 0.5 & 0.0 & 0.4 \\
    \bottomrule
  \end{tabular}
  \caption{\textbf{Detailed comparison} on the 100-part benchmark, tolerance $\pm0.1$\,mm, probe depth cap 20\,mm. Values are means over the successfully scored parts (99 for the GPT-6 Astra agent and CNC-Net, 100 for MillBill and the Opus 5.5 agent, 82 for the Qwen3.8-27B agent); failures are reported separately without imputed scores (GPT-6 Astra agent: evaluation failure on \texttt{p02203}; Qwen3.8-27B agent: G-code generation errors on 18 parts, such as incorrect arcs and drilling cycles and skipped operations; CNC-Net: no output geometry for \texttt{p04119}). Volume proxies are normal-depth integrals, not exact volumes. Maximum depths are means of per-part maxima. MillBill is evaluated with the tessellated stock; the code-writing agents with analytic stock primitives where available and a mesh fallback otherwise; CNC-Net with its native bounding-box stock and the saved boundary-corrected export. Within-tolerance scores excluding expected deviations equal the reported ones, because no expected deviations are supplied.}
  \label{tab:suppl:geometry}
\end{table}

\section{Additional results}
\label{sec:suppl:results}

\Cref{tab:suppl:tolerance} scores the same programs at other tolerances (\cref{sec:exp:analysis}).

\begin{table}[t]
  \centering
  \small
  \setlength{\tabcolsep}{4pt}
  \begin{tabular}{@{}lccccc@{}}
    \toprule
    Method & Scored & $\tau = 0.05$ & $0.1$ & $0.2$ & $0.5$ \\
    \midrule
    CNC-Net & 99/100 & 38.3 & 42.3 & 47.7 & 57.3 \\
    GPT-6 Astra & 99/100 & \textbf{90.2} & 94.3 & 95.3 & 95.6 \\
    Opus 5.5 & 100/100 & 78.9 & 83.0 & 86.0 & 88.2 \\
    \midrule
    MillBill & 100/100 & 89.3 & \textbf{97.4} & \textbf{98.0} & \textbf{98.1} \\
    \bottomrule
  \end{tabular}
  \caption{\textbf{Sensitivity to the tolerance.} $\text{Acc}_\tau$ (\%) of the same programs scored at different tolerances $\tau$ (mm), averaged over the scored parts.}
  \label{tab:suppl:tolerance}
\end{table}

\begin{figure*}[t]
  \centering
  \includegraphics[width=0.8\linewidth]{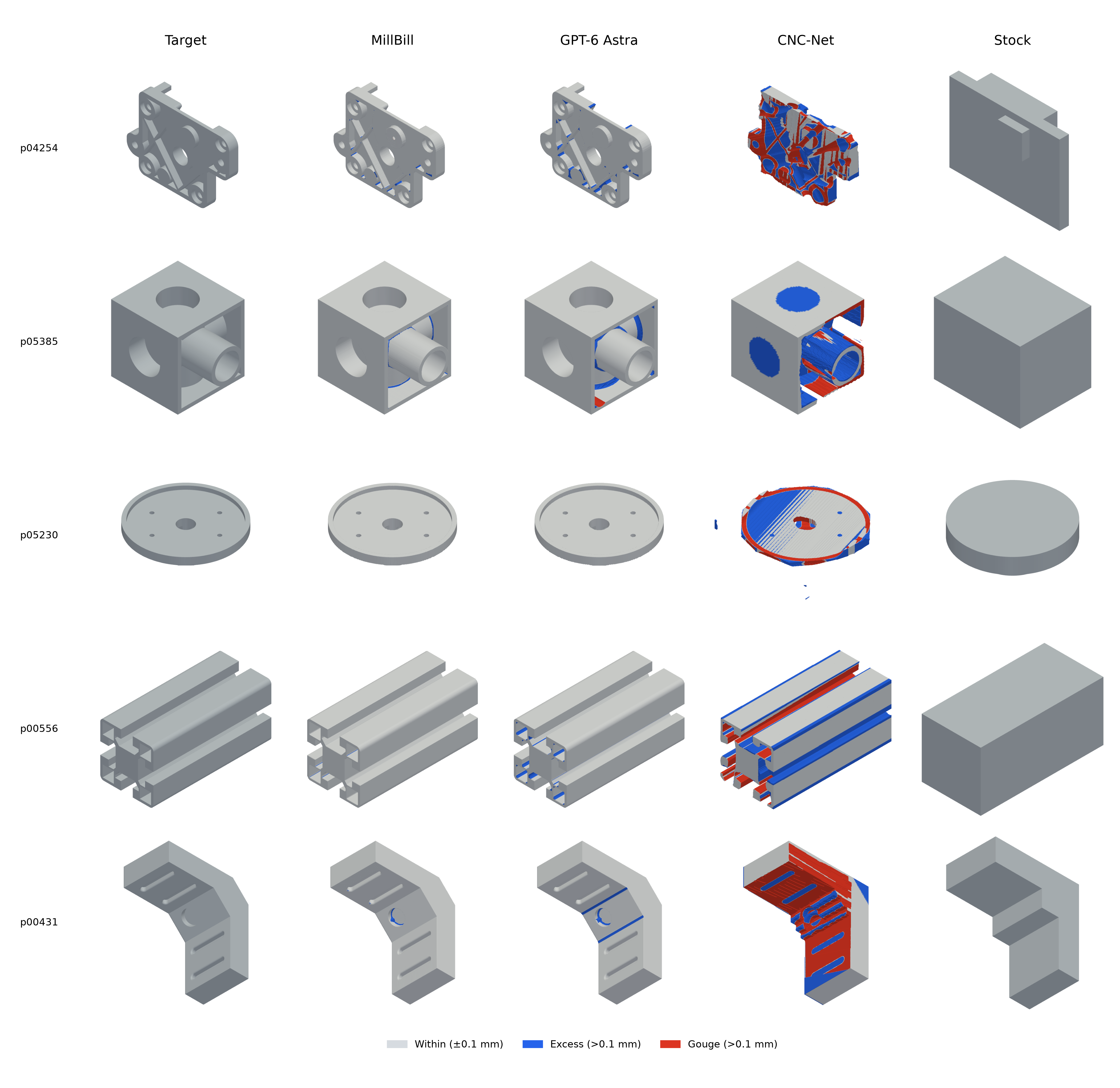}
  \caption{\textbf{Further simulated results.} Five parts not shown in \cref{fig:qualitative}, laid out and coloured as there.}
  \label{fig:suppl:gallery}
\end{figure*}

\Cref{tab:suppl:perpart} lists the results per part, and \cref{fig:suppl:gallery} shows the simulated result of every method on further parts.

\begin{table*}[p]
  \centering
  \footnotesize
  \setlength{\tabcolsep}{4pt}
  \begin{tabular}{@{}lcccccc@{\hspace{2em}}lcccccc@{}}
    \toprule
    Part & Setups & CNC-Net & Qwen & Opus & Astra & MillBill & Part & Setups & CNC-Net & Qwen & Opus & Astra & MillBill \\
    \midrule
    p00089 & 6 & 48.6 & -- & 89.2 & 87.9 & 95.2 & p03332 & 6 & 59.1 & 4.9 & 71.9 & 88.6 & 91.0 \\
    p00114 & 4 & 58.6 & 0.3 & 95.1 & 95.1 & 93.5 & p03369 & 4 & 55.5 & -- & 94.8 & 97.6 & 98.2 \\
    p00178 & 6 & 15.8 & -- & 83.6 & 82.3 & 90.0 & p03383 & 4 & 50.6 & 0.1 & 48.2 & 100.0 & 99.8 \\
    p00431 & 6 & 58.1 & 0.0 & 94.3 & 98.5 & 99.4 & p03500 & 4 & 27.1 & -- & 98.3 & 100.0 & 99.7 \\
    p00445 & 1 & 47.7 & 50.0 & 97.2 & 99.3 & 96.6 & p03502 & 4 & 42.7 & 0.1 & 98.8 & 98.8 & 99.8 \\
    p00486 & 4 & 50.0 & 0.2 & 89.7 & 100.0 & 99.2 & p03548 & 6 & 20.8 & 0.0 & 73.6 & 99.2 & 97.8 \\
    p00531 & 6 & 7.5 & 0.0 & 87.8 & 1.1 & 94.2 & p03553 & 4 & 31.3 & 0.0 & 91.7 & 99.9 & 95.2 \\
    p00556 & 4 & 27.9 & 0.1 & 52.8 & 83.1 & 97.7 & p03674 & 3 & 26.4 & -- & 93.8 & 95.7 & 98.0 \\
    p00696 & 6 & 39.3 & 0.5 & 100.0 & 100.0 & 99.8 & p03718 & 4 & 49.6 & 0.0 & 39.9 & 63.2 & 96.5 \\
    p00702 & 6 & 64.7 & 0.0 & 98.9 & 100.0 & 99.5 & p03881 & 4 & 52.5 & 0.0 & 95.8 & 98.7 & 99.3 \\
    p00741 & 6 & 31.8 & 0.0 & 42.2 & 93.4 & 98.6 & p03901 & 5 & 41.5 & 6.3 & 95.1 & 96.6 & 93.5 \\
    p00749 & 6 & 57.2 & 0.0 & 98.3 & 99.4 & 99.4 & p03907 & 4 & 29.0 & 1.0 & 71.7 & 100.0 & 99.5 \\
    p00770 & 6 & 12.1 & 0.1 & 40.4 & 99.9 & 96.5 & p04119 & 4 & -- & 0.0 & 76.6 & 98.7 & 98.6 \\
    p00820 & 4 & 16.3 & 0.1 & 84.1 & 89.7 & 98.9 & p04164 & 4 & 10.9 & 0.3 & 91.7 & 88.0 & 97.1 \\
    p00938 & 3 & 64.7 & 0.0 & 100.0 & 100.0 & 99.5 & p04237 & 6 & 2.4 & 0.0 & 57.5 & 97.1 & 99.2 \\
    p01083 & 4 & 43.1 & 23.5 & 96.1 & 98.4 & 98.5 & p04254 & 6 & 23.4 & 3.4 & 76.5 & 93.5 & 95.8 \\
    p01166 & 4 & 46.2 & 0.0 & 81.1 & 98.7 & 98.8 & p04285 & 4 & 61.6 & -- & 90.1 & 91.5 & 89.3 \\
    p01263 & 6 & 47.9 & 0.1 & 97.9 & 100.0 & 99.0 & p04300 & 4 & 78.5 & 3.3 & 100.0 & 100.0 & 99.4 \\
    p01518 & 3 & 49.6 & 0.0 & 92.8 & 91.7 & 96.7 & p04382 & 4 & 53.0 & -- & 80.3 & 80.3 & 79.3 \\
    p01558 & 3 & 56.0 & 0.0 & 98.7 & 99.7 & 98.9 & p04438 & 3 & 50.5 & 0.0 & 82.8 & 90.9 & 98.8 \\
    p01582 & 5 & 85.3 & 72.2 & 99.9 & 100.0 & 98.0 & p04456 & 4 & 7.4 & -- & 65.2 & 77.3 & 98.9 \\
    p01703 & 1 & 74.8 & 46.6 & 88.2 & 99.3 & 86.5 & p04462 & 6 & 31.6 & 2.3 & 90.6 & 73.3 & 97.2 \\
    p01816 & 4 & 85.6 & 0.0 & 99.4 & 100.0 & 99.9 & p04650 & 1 & 75.5 & 3.3 & 98.7 & 98.7 & 98.3 \\
    p01870 & 4 & 25.2 & 7.5 & 100.0 & 100.0 & 99.7 & p04658 & 5 & 21.3 & 0.0 & 66.9 & 88.6 & 99.8 \\
    p01872 & 3 & 50.5 & 0.0 & 100.0 & 100.0 & 98.6 & p04676 & 6 & 6.8 & 0.0 & 65.0 & 67.1 & 98.2 \\
    p01958 & 6 & 23.2 & -- & 68.7 & 78.9 & 99.2 & p04689 & 6 & 27.2 & 0.0 & 45.6 & 98.8 & 99.1 \\
    p02058 & 4 & 8.5 & 0.0 & 98.4 & 98.4 & 98.7 & p04698 & 4 & 40.1 & 0.0 & 17.5 & 100.0 & 99.8 \\
    p02092 & 6 & 34.7 & 1.4 & 77.7 & 91.8 & 98.0 & p04699 & 6 & 9.2 & -- & 95.5 & 99.6 & 99.6 \\
    p02146 & 1 & 88.5 & 70.0 & 100.0 & 100.0 & 98.4 & p04710 & 5 & 39.1 & 23.0 & 100.0 & 100.0 & 99.7 \\
    p02191 & 5 & 5.3 & 0.0 & 64.2 & 84.8 & 96.6 & p04726 & 4 & 58.2 & 0.1 & 99.5 & 100.0 & 99.9 \\
    p02203 & 4 & 7.9 & -- & 49.1 & -- & 99.8 & p04781 & 4 & 52.2 & 0.8 & 85.7 & 99.9 & 97.9 \\
    p02207 & 6 & 27.2 & 10.2 & 91.6 & 94.3 & 98.8 & p04783 & 4 & 47.4 & 0.1 & 93.9 & 99.6 & 98.3 \\
    p02274 & 3 & 78.3 & 0.3 & 96.0 & 93.0 & 99.0 & p04821 & 4 & 31.9 & 0.0 & 100.0 & 100.0 & 99.5 \\
    p02332 & 6 & 32.1 & 0.0 & 86.4 & 99.9 & 99.9 & p04831 & 4 & 11.8 & 0.1 & 100.0 & 100.0 & 99.5 \\
    p02334 & 1 & 89.3 & 39.2 & 83.3 & 100.0 & 99.9 & p04843 & 3 & 69.3 & 0.0 & 96.1 & 99.9 & 98.5 \\
    p02418 & 5 & 31.9 & -- & 84.1 & 98.0 & 99.5 & p04870 & 6 & 11.3 & 3.1 & 89.5 & 58.4 & 98.8 \\
    p02687 & 4 & 93.5 & 21.7 & 100.0 & 100.0 & 99.5 & p04991 & 2 & 93.1 & -- & 100.0 & 100.0 & 99.0 \\
    p02817 & 6 & 24.6 & 0.1 & 46.9 & 98.3 & 97.3 & p05005 & 4 & 14.1 & -- & 94.6 & 100.0 & 99.4 \\
    p02837 & 1 & 82.0 & 0.0 & 100.0 & 100.0 & 98.7 & p05025 & 6 & 9.1 & -- & 68.0 & 99.0 & 98.4 \\
    p02853 & 1 & 77.7 & 35.3 & 91.2 & 98.3 & 97.2 & p05053 & 5 & 8.8 & 0.2 & 88.5 & 99.0 & 99.3 \\
    p02863 & 5 & 23.8 & 6.7 & 70.3 & 98.6 & 98.4 & p05088 & 1 & 85.9 & 0.7 & 96.1 & 93.5 & 78.3 \\
    p02905 & 1 & 76.2 & 64.4 & 88.9 & 100.0 & 100.0 & p05230 & 5 & 31.2 & 0.1 & 58.2 & 94.6 & 98.9 \\
    p03081 & 3 & 5.6 & -- & 76.7 & 95.6 & 96.5 & p05385 & 6 & 23.3 & 0.2 & 59.6 & 88.9 & 95.7 \\
    p03086 & 4 & 10.4 & -- & 100.0 & 100.0 & 99.8 & p05427 & 1 & 99.7 & 91.9 & 99.9 & 100.0 & 99.9 \\
    p03152 & 6 & 69.4 & 0.1 & 37.9 & 99.7 & 99.8 & p05451 & 2 & 99.1 & 0.0 & 63.8 & 99.3 & 64.5 \\
    p03181 & 6 & 61.8 & 1.2 & 95.3 & 100.0 & 99.0 & p05541 & 2 & 15.5 & 23.8 & 53.5 & 97.6 & 98.6 \\
    p03189 & 5 & 27.2 & -- & 88.0 & 86.9 & 98.7 & p05597 & 4 & 31.6 & 0.3 & 88.8 & 99.6 & 99.8 \\
    p03209 & 6 & 21.9 & 0.5 & 48.7 & 99.8 & 99.8 & p05687 & 6 & 29.6 & 0.2 & 98.0 & 99.2 & 98.7 \\
    p03218 & 6 & 25.8 & 5.8 & 100.0 & 100.0 & 99.8 & p05719 & 6 & 45.5 & 13.4 & 91.4 & 97.5 & 96.3 \\
    p03242 & 6 & 16.9 & 3.7 & 52.3 & 96.7 & 99.3 & p05791 & 3 & 83.2 & 0.0 & 99.4 & 99.9 & 99.7 \\
    \bottomrule
  \end{tabular}
  \caption{$\text{Acc}_\tau$ (\%) per part at $\tau = 0.1$\,mm. Qwen, Opus and Astra: the code-writing agents built on Qwen3.8-27B, Claude Opus 5.5 and GPT-6 Astra. --: not scored (\cref{tab:suppl:geometry}). Setups: setups executed by MillBill, including failed and discarded ones.}
  \label{tab:suppl:perpart}
\end{table*}

\begin{figure}[t]
  \centering
  \includegraphics[width=0.75\linewidth]{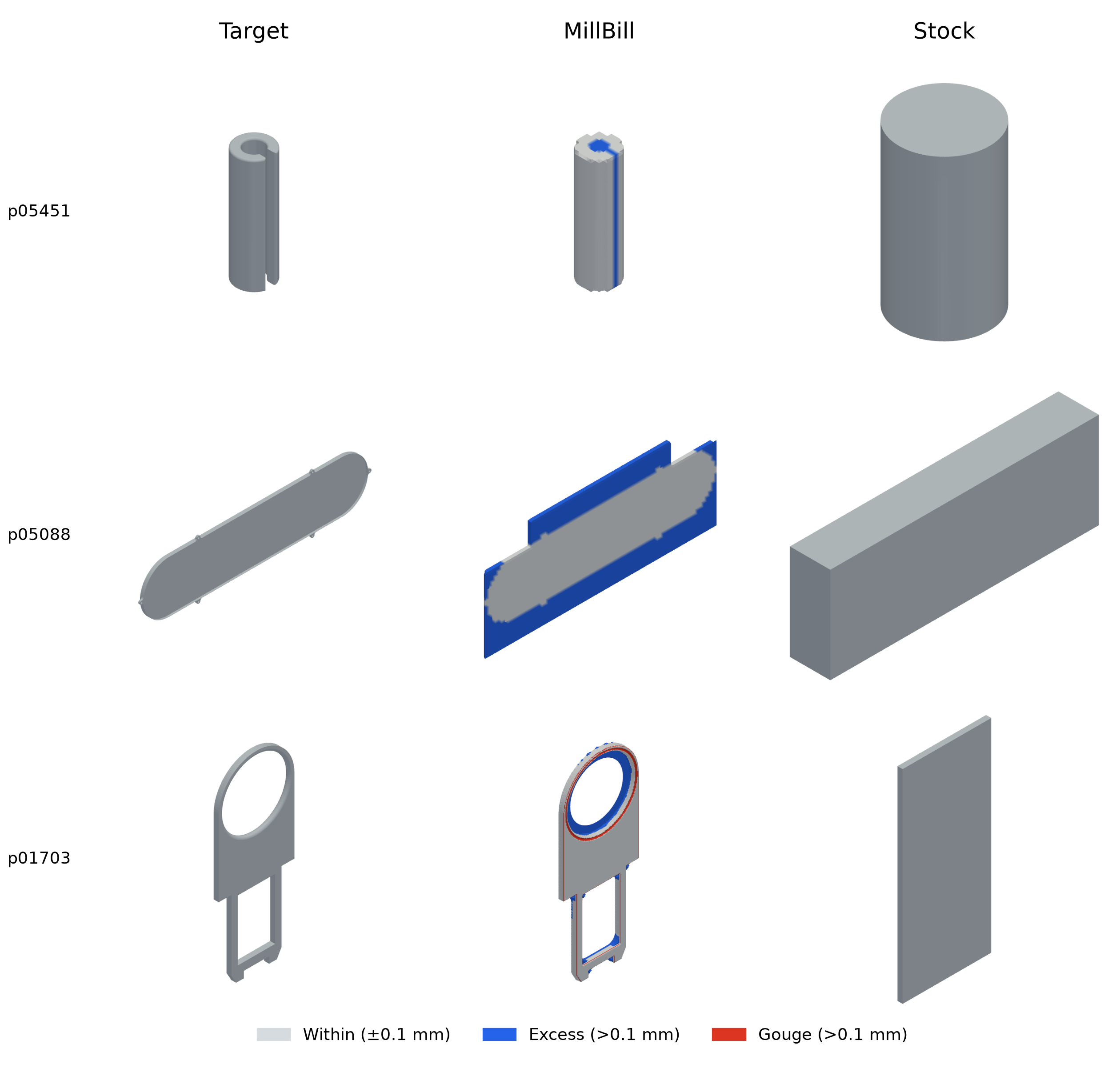}
  \caption{\textbf{Failure cases.} Target, MillBill's simulated workpiece and the stock for three of the parts on which MillBill scores lowest ($\text{Acc}_{0.1}$ of 64.5\%, 78.3\% and 86.5\%, \cref{tab:suppl:perpart}), coloured as in \cref{fig:qualitative}.}
  \label{fig:suppl:failures}
\end{figure}

\paragraph{Example plan and feedback.}
Part p02334 is machined in a single setup. Its recognised features are pockets, a slot, fillets, an outer profile and a cavity, and the initial plan consists of facing, cavity roughing, fillet milling and deburring. After the first build the simulation leaves 92.46\% of the surface within tolerance and 7.51\% gouged. The model removes the deburring operation; the second build reaches 99.92\% within tolerance with no gouged area and is kept as the final program. The fillet milling was skipped because the harness has no CAM mapping for it, so this example shows the geometric improvement rather than the full execution of the intended plan.

\end{document}